\documentclass[journal]{IEEEtran}

\usepackage[T1]{fontenc}
\usepackage{dfadobe}  

\usepackage{times}
\usepackage{epsfig}
\usepackage{graphicx}
\usepackage{amsmath}
\usepackage{amssymb}
\usepackage{bbold}
\usepackage{booktabs}

\usepackage{multirow}
\usepackage[pagebackref=true,breaklinks=true,letterpaper=true,colorlinks,bookmarks=false]{hyperref}

\usepackage[dvipsnames]{xcolor}
\usepackage[normalem]{ulem}
\definecolor{mred}{rgb}{.80,.12,.30}
\definecolor{MRED}{rgb}{.80,.12,.30}
\definecolor{grey}{rgb}{0.5,0.5,0.5}
\definecolor{lgrey}{rgb}{0.7,0.7,0.7}
\definecolor{purple}{rgb}{.75,0,.85}
\definecolor{cyan}{rgb}{0,0.68,.94}
\definecolor{pistachio}{rgb}{0.58, 0.77, 0.45}
\definecolor{myorange}{rgb}{0.94, 0.36, 0.13}

\newif\ifnotes
\notestrue

\newcommand{\ra}[1]{\renewcommand{\arraystretch}{#1}}

\begin{document}

\newcommand{\eg}{\emph{e.g.}}
\newcommand{\ie}{\emph{i.e.}}

\title{FloorLevel-Net: Recognizing Floor-Level Lines with Height-Attention-Guided Multi-task Learning}

\author{Mengyang~Wu, Wei~Zeng,~\IEEEmembership{Member,~IEEE} and~Chi-Wing~Fu,~\IEEEmembership{Member,~IEEE}% <-this % stops a space
\thanks{M. Wu, and C.-W. Fu are with the Chinese University of Hong Kong. e-mail: \{mywu, cwfu\}@cse.cuhk.edu.hk.}% <-this % stops a space
\thanks{W. Zeng is with Shenzhen Institute of Advanced Technology, Chinese Academy of Sciences. e-mail: wei.zeng@siat.ac.cn.}% <-this % stops a space

\thanks{W. Zeng and C.-W. Fu are the corresponding authors.}
\thanks{This paper has supplementary downloadable material available at http://ieeexplore.ieee.org., provided by the author. The material includes additional results and comparisons. Contact mywu@cse.cuhk.edu.hk for further questions about this work.}
}

\maketitle

\begin{abstract}
The ability to recognize the position and order of the floor-level lines that divide adjacent building floors can benefit many applications, for example, urban augmented reality (AR).
This work tackles the problem of locating floor-level lines in street-view images, using a supervised deep learning approach.
Unfortunately, very little data is available for training such a network $-$ current street-view datasets contain either semantic annotations that lack geometric attributes, or rectified facades without perspective priors. 
To address this issue, we first compile a new dataset and develop a new data augmentation scheme to synthesize training samples by harassing  (i) the rich semantics of existing rectified facades and (ii) perspective priors of buildings in diverse street views.
Next, we design \emph{FloorLevel-Net}, a multi-task learning network that associates explicit features of building facades and implicit floor-level lines, along with a height-attention mechanism to help enforce a vertical ordering of floor-level lines.
The generated segmentations are then passed to a second-stage geometry post-processing to exploit self-constrained geometric priors for plausible and consistent reconstruction of floor-level lines.
Quantitative and qualitative evaluations conducted on assorted facades in existing datasets and street views from Google demonstrate the effectiveness of our approach.
Also, we present context-aware image overlay results and show the potentials of our approach in enriching AR-related applications.
{\textit{Project website: \url{https://wumengyangok.github.io/Project/FloorLevelNet}.}}

\end{abstract}


%%%%%%%%%%%%%%%%%%%%%%%%%%%%%%%%%%%%%%%%%%%%%%%%%%%%%%%%%%%%%%%%%%%%%%%%%
\section{Introduction}
\label{sec:intro}

\begin{figure}[t]
\centering
\includegraphics[width=0.99\linewidth]{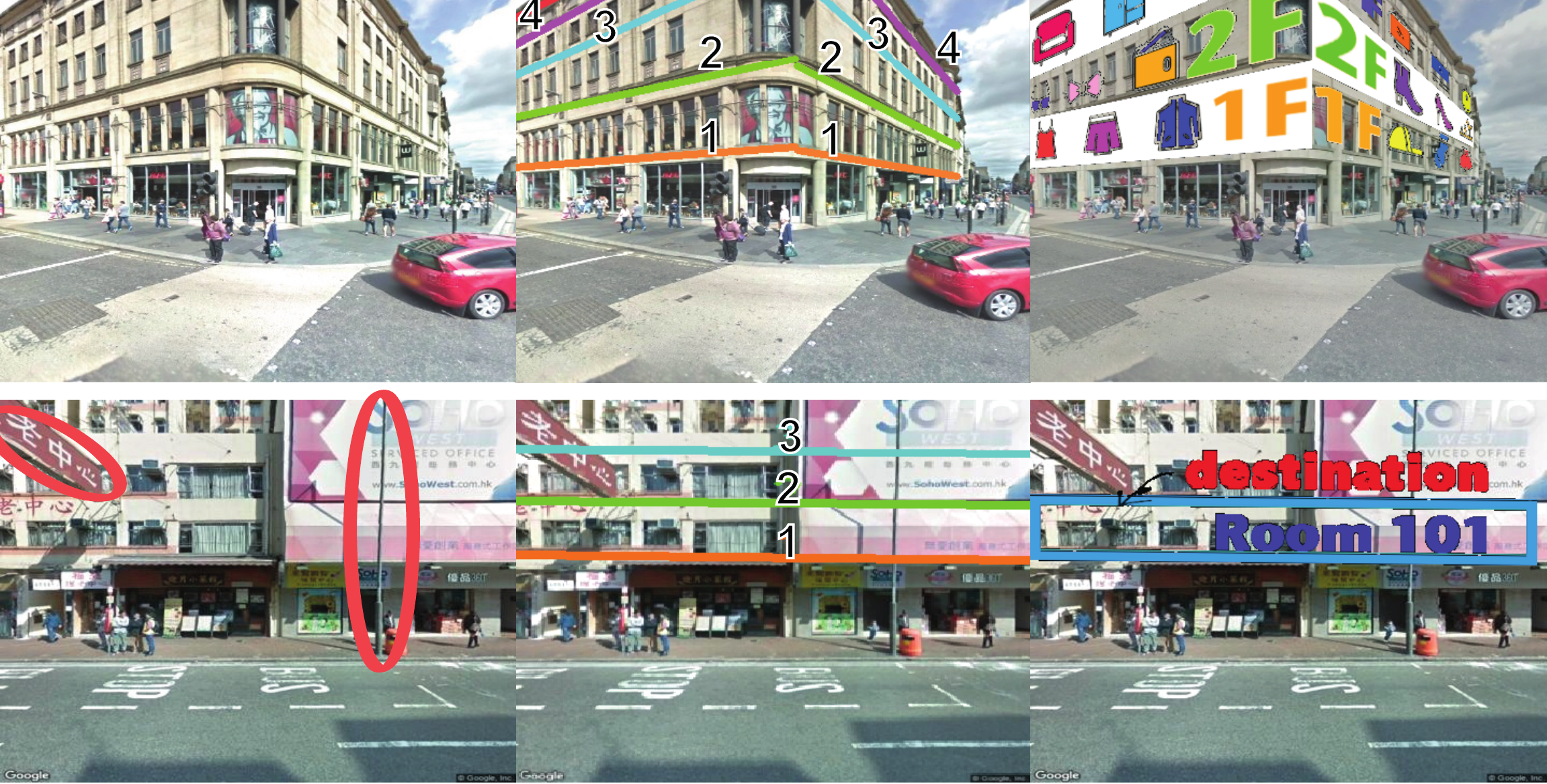}
\vspace{-5.5mm}
\caption{\textbf{Left} column shows two example street-view images in London (top) and Hong Kong (bottom), where the camera views are \textit{side}- and \emph{front}-facing relative to the building, respectively. Note the occlusions introduced by the advertisement billboard and light post circled in red on bottom left.
\textbf{Middle} column shows floor-level lines recognized by our method with geometric positions and semantic order labels.
\textbf{Right} column shows potential floor-aware image-overlay results to aid shopping (top) and navigation (bottom).
}
\label{fig:floor-lines}
\vspace{-2mm}
\end{figure}

\IEEEPARstart{F}{loor-level} lines are line segments that separate {\em adjacent floors\/} on a building facade; see Fig.~\ref{fig:floor-lines} (middle).
Being able to recognize them in city-wide street views can benefit various applications,~\emph{e.g.}, urban 3D reconstruction~\cite{musialski_2013_survey}, building topology analysis~\cite{wu_2018_city-wide}, and urban vitality study~\cite{zeng_2018_vitalvizor}.
Intrinsically, floor-level lines for the same building are parallel, through which we can reconstruct the homography of the facade in a perspective view~\cite{criminisi_2000_single} and support floor-aware augmented reality (AR) applications; see Fig.~\ref{fig:floor-lines} (right).

This work considers the problem of inferring floor-level lines in street-view images, which requires the recovery of not only \textit{geometric} priors (\emph{e.g.}, positions and vanishing directions) in the image view, but also \textit{semantic} information (\emph{e.g.}, floor orders) relevant to the floor-level lines.
The task is relatively intuitive for humans but very challenging for computers.
So far, we are not aware of any work that can robustly detect and recognize floor-level lines.

Related prior works on 3D reconstruction (\emph{e.g.},~\cite{barinova_2008_fast, gupta_2010_blocks}), and facade parsing (\emph{e.g.},~\cite{martinovic_2012_three-layered, yang_2012_parsing}), typically rely on various assumptions about the structural regularity on building facades,~\emph{e.g.}, repetitive windows and balconies (\eg,~\cite{schindler_2008_detecting, wu_2011_repetition}).
Though these extra constraints help disambiguate the problem and the results might help infer floor-level lines, they still have various limitations, such as being error-prone when handling perspective-oriented {facades} (Fig.~\ref{fig:floor-lines} top-left) and scenes with occlusions (Fig.~\ref{fig:floor-lines} bottom-left).
Alternatively, we may infer floor-level lines by locating line segments roughly in the same direction.
However, existing line detectors (\emph{e.g.},~\cite{lsd_2012, gilles_2018_contrario}) can easily generate huge amount of irrelevant line segments in cluttered street views, and offer no semantic information for inferring the floor order.
Recently, Lee et al.~\cite{lee_2017_semantic} proposed the concept of semantic lines that characterize the spatial scene structure in images.
The method first finds candidate lines using line features at multi-scale pooling layers, then filters out semantic lines using local line features.
However, the filtering process (semantic or not) is a binary classification problem, whilst this work requires to recover not only a multi-class label per line but also a plausible floor order for lines in the same {facade},~\emph{i.e.}, order 1, order 2,~\emph{etc}.

Recent attempts of using deep neural networks for semantic scene segmentation (\eg,~\cite{Choi_2020_HANet}) and planar surface recognition (\emph{e.g.},~\cite{yang_2018_recovering, zhiliang_2020_arscape}) exploit the possibility of jointly learning the semantic and geometric attributes in street views. 
The methods achieve superior performances over previous deep learning methods that use solely the semantic features. 
This work also leverages a supervised deep learning approach that can jointly infer geometric priors and semantic information of floor-level lines.
This is nevertheless a challenging task.
First, a key requirement for network training is the availability of large, labeled street-view images with annotations of floor-level lines.
However, existing street-view datasets contain either semantic annotations that lack geometric attributes, or rectified facades without perspective priors. 
To address this issue, we devise a new data augmentation scheme and compile a new dataset by effectively combining the rich context of floor-level lines marked on an existing facade dataset, with perspective priors easily-extracted on street-view images.
In this way, we can largely reduce the manual workload in the dataset construction, while promoting the generalizability of our network model.

Second, we are not aware of any network architecture that can fulfill the requirement of recognizing and locating floor-level lines in street-view images.
To fill the gap, we design \emph{FloorLevel-Net} $-$ a new deep learning framework with two main components inspired by the characteristics of our floor-level dataset.
(i) \emph{FloorLevel-Net} leverages multi-task learning that associates geometric properties of {facades} orientation to camera, with semantic features of facade appearance, including windows, doors, and balconies, etc.
(ii) \emph{FloorLevel-Net} incorporates a height-attention mechanism to enforce vertical orderings of floor-level lines, as floor order naturally increases from bottom to top in image space.
We further exploit various geometric properties, including the \emph{vanishing point (VP)} and \emph{floor order} constraints, to enforce a consistent reconstruction of floor-level lines from piecewise segmentations by \emph{FloorLevel-Net}.

We evaluate the performance of our approach on both building images from existing facade datasets and self-collected street-view images from Google Street View (GSV)~\cite{gsv}.
Both qualitative and quantitative analysis demonstrate the effectiveness of the individual components in our approach.
The main contributions of this work include:

\begin{itemize}

\item 
We propose a \emph{data augmentation} scheme to integrate rich semantics of rectified facades and geometric priors of {buildings} in diverse street views, and compile a new dataset for recognizing floor-level lines (Sec.~\ref{sec:data_composition}).

\vspace{1mm}
\item 
We design \emph{FloorLevel-Net}$-$a \emph{multi-task CNN architecture with height attention} to simultaneously predict a facade segmentation map and a floor-level distribution map for an input street-view image (Sec.~\ref{ssec:level_feature} to~\ref{ssec:implementation}).
We further develop a post-processing framework to extract floor-level parameters (including the endpoints and vanishing points), and refine the parameters with \emph{self-constrained geometric priors} (Sec.~\ref{ssec:line_params}).

\vspace{1mm}
\item 
We evaluate the effectiveness of our approach in recognizing floor-level lines, and demonstrate the applicability in enriching \textit{context-aware urban AR} (Sec.~\ref{sec:experiment}).

\end{itemize}
% \vspace*{-1mm}
\section{Related Work}
\label{sec:related_work}

The recognition of geometric structures in urban scenes, \eg, surface layout~\cite{derek_2007_recovering} and driving lanes~\cite{zhang_2018_geometric}, has been gaining attention in image processing and vision research.
This work targets at floor-level lines that separate adjacent floors on building facades. 
A closely-related topic is to reconstruct 3D building models from a monocular image.
Conventional approaches typically adopt a two-stage approach: first parse a facade into piecewise regions like windows, balconies, etc.~\cite{martinovic_2012_three-layered, yang_2012_parsing}, then split the regions into regular layouts for subsequent modeling,~\emph{e.g.}, by shape grammars~\cite{mueller_2007_image-based, teboul_2011_rl}. 
{Existing methods, however, rely heavily on detecting repetition~\cite{wu_2011_repetition, schindler_2008_detecting}, symmetric or rectangular structures on facade layout~\cite{teboul_2010_shape, weissenber_2013_there}, thus exhibiting significant challenges for parsing general urban scenes in the wild.}
Particular obstacles include perspective-oriented {facades} (\eg, see Fig.~\ref{fig:floor-lines} (top) and the red inset in Fig.~\ref{fig:overview}) and scene occlusions by billboards and traffic lights (\eg, see Fig.~\ref{fig:floor-lines} (bottom) and the blue inset in Fig.~\ref{fig:overview}), resulting in failures in matching repetition/symmetry/rectangular constraints.
It is hard to identify common geometric assumptions that can well fit diverse styles of building {facades} in different cities.

Recent studies on urban scene recognition focus on deep-learning-based approaches, benefiting from new datasets of urban scenes and advancements in neural network architecture, \eg,~\cite{Long_2015_FCN, Zhao_2017_PSP, Chen_2018_Deeplab}.
Specifically, DeepFacade~\cite{liu_2017_deep_facade, liu_2020_deep_facade} parses {facades} using a fully convolutional network to produce pixel-wise semantics.
Some work also attempts to formulate geometric understanding of urban scenes as an image segmentation problem.
For instance, Haines and Calway~\cite{haines_2015_recognising} infer surface orientations using spatiograms of gradients and colors.
However, the results are piecewise segments that are typically discontinuous with coarse and irregular boundaries, hindering the recognition of geometric priors like the endpoints of floor-level lines.
To further recover geometric priors, a second-stage post-processing framework can be employed.
Zeng et al.~\cite{zhiliang_2020_arscape} employ a vanishing-point-constrained optimization to enhance piecewise segmentations of planar building facades.

Similarly, we adopt a twofold process to recognize floor-level lines on building {facades}.
First, we predict a segmentation mask of pixel-wise line labels (floor-level distribution) using a multi-task CNN architecture.
Second, we refine the piecewise floor-level distribution into line parameters using self-constrained geometric priors.
Nevertheless, the work faces severe challenges of being lack of \emph{training dataset} and \emph{suitable network architecture}.
We tackle the challenges from the following perspectives:

\begin{itemize}

\vspace{1mm}
\item
\textbf{Data augmentation} is a vital technique to improve the diversity of the training data and the network generalizability.
There are several publicly available datasets for urban scene (\eg, KITTI~\cite{geiger_2012_are} and Cityscapes~\cite{cordts_2016_cityscapes}) and facade (\eg, CMP~\cite{Tylecek13} and LabelMeFacade~\cite{Froehlich-Rodner-Denzler-ICPR2010}).
Yet, we cannot use them directly in this work.
In contrast to explicit building structures such as windows and balconies that have obvious visual appearance, floor-level lines are rather implicit, requiring contextual information for the recognition.
To fill the gap, we propose a new data augmentation scheme that leverages detailed facade semantics of front-facing {facades} in CMP~\cite{Tylecek13} and diverse building perspectives in GSV images.
In this way, the synthesized dataset efficiently captures both the rich context of floor-level lines and geometric priors of building {facades} in reality.

\begin{figure*}[ht]
\centering
\includegraphics[width=0.99\linewidth]{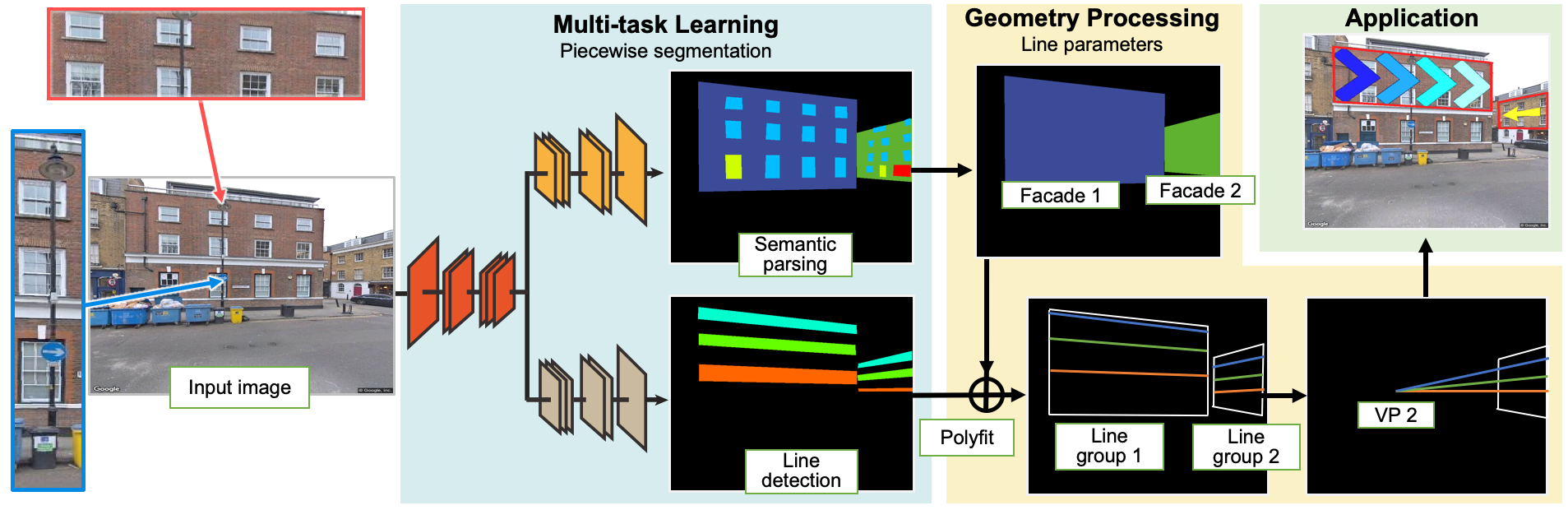}
\vspace{-3mm}
\caption{
Our two-stage approach:
(i) \emph{FloorLevel-Net} is a multi-task learning network that segments the input image into building-facade-wise semantic regions (top) and floor-level distributions (bottom); and
(ii) our method further fits and refines the pixel-wise network outputs into polylines with geometric parameters.
Further, we can take the reconstructed floor-level lines to support and enrich urban AR applications with floor-aware image overlay.
}
\label{fig:overview}
\vspace{-3mm}
\end{figure*}

\vspace{2mm}
\item
\textbf{Multi-task learning} helps to boost the performance of many deep-learning-based vision tasks, by learning features from relevant knowledge domains.
In the literature, various efforts have been devoted to exploit multi-task learning for geometric understanding of urban scenes.
For example, Liu et al.~\cite{liu_2017_single} tackle scene recognition and reconstruction tasks in a tightly-coupled framework; besides, there are assorted works for indoor scene understanding (\eg, \cite{Wang_2015_NormalEdge, Ren_2018_CrossDomain, Eigen_2015_DepthNormal}) by fusing features of depth, surface normals, and objects.
Human experience on locating floor-level lines typically relies on recognizing the {facades} and its enclosing regions, \eg, doors and windows.
Therefore, we divide our task into subtasks of recognizing {facades} and floor-level lines, and design a multi-task learning network to simultaneously address the two subtasks together.

\vspace{2mm}
\item
\textbf{Channel-wise attention} exploits the inter-channel relationship of features and scales the feature map according to the importance of each channel. The mechanism is first proposed in SENet~\cite{hu_2018_SEN} and adopted widely in image classification (\eg,~\cite{Woo_2018_CBAM}) and segmentation (\eg,~\cite{Yu_2018_Learning}).
Recent attention approaches take more advantages of contextual information inside the image domains,~\eg, the criss-cross attention module by Huang et al.~\cite{huang_2019_ccnet} and recursive context routing (ReCoR) mechanism by Chen et al.~\cite{chen_2021_recursive}.
Choi et al.~\cite{Choi_2020_HANet} utilize the unbalanced class distribution at varying vertical locations in urban scene images, and design a height-attention module to emphasize the unbalance.
Floor-level lines in our dataset exhibit a similar property, since floor order on the same {facade} always increases from bottom to top in the image space.
This drives us to adopt a similar configuration in the recognition of floor-level lines.

\end{itemize}
\section{Overview}
\label{sec:overview}

\noindent
In this work, our goal is
\begin{quote}
{\em to recognize and locate the floor-level lines on each building {facade} (that is close to the camera) in the image space of street-view images.
\/}
\end{quote}
The input to our method is a single RGB image $\mathcal{I} \in \mathbb{R}^{W \times H \times 3}$ of width $W$ and height $H$.
Given $\mathcal{I}$, we aim to predict a set of line segments $\mathcal{L} = \{L_1,...,L_n\}$ per {facade} that marks the separations between adjacent floors; see Fig.~\ref{fig:floor-lines} (middle) for two examples.
In the following, we refer {\em line segments\/} as lines for simplicity.
Each line $L_i$ contains {\em a pair of endpoints\/} ($x_s^i,\, y_s^i$) and ($x_e^i,\, y_e^i$) in the image space of $\mathcal{I}$.
Also, we aim to recognize a {\em floor order\/} (denoted as $l^i \in \{ 1, 2, 3, ...\}$) per line $L_i$, such that the line that separates the ground floor and the first floor has a floor order value of one, the next line above has a floor order value of two, etc.
So, each $L_i$ is specified as a 5-tuple $\{x^{i}_{s},\, y^{i}_{s},\, x^{i}_{e},\, y^{i}_{e},\, l^{i}\}$.
As a visualization, we use the same coloring scheme (orange for order 1, green for order 2, etc.) to reveal the floor orders in all our results.

\begin{figure*}[ht]
\centering
\includegraphics[width=0.92\linewidth]{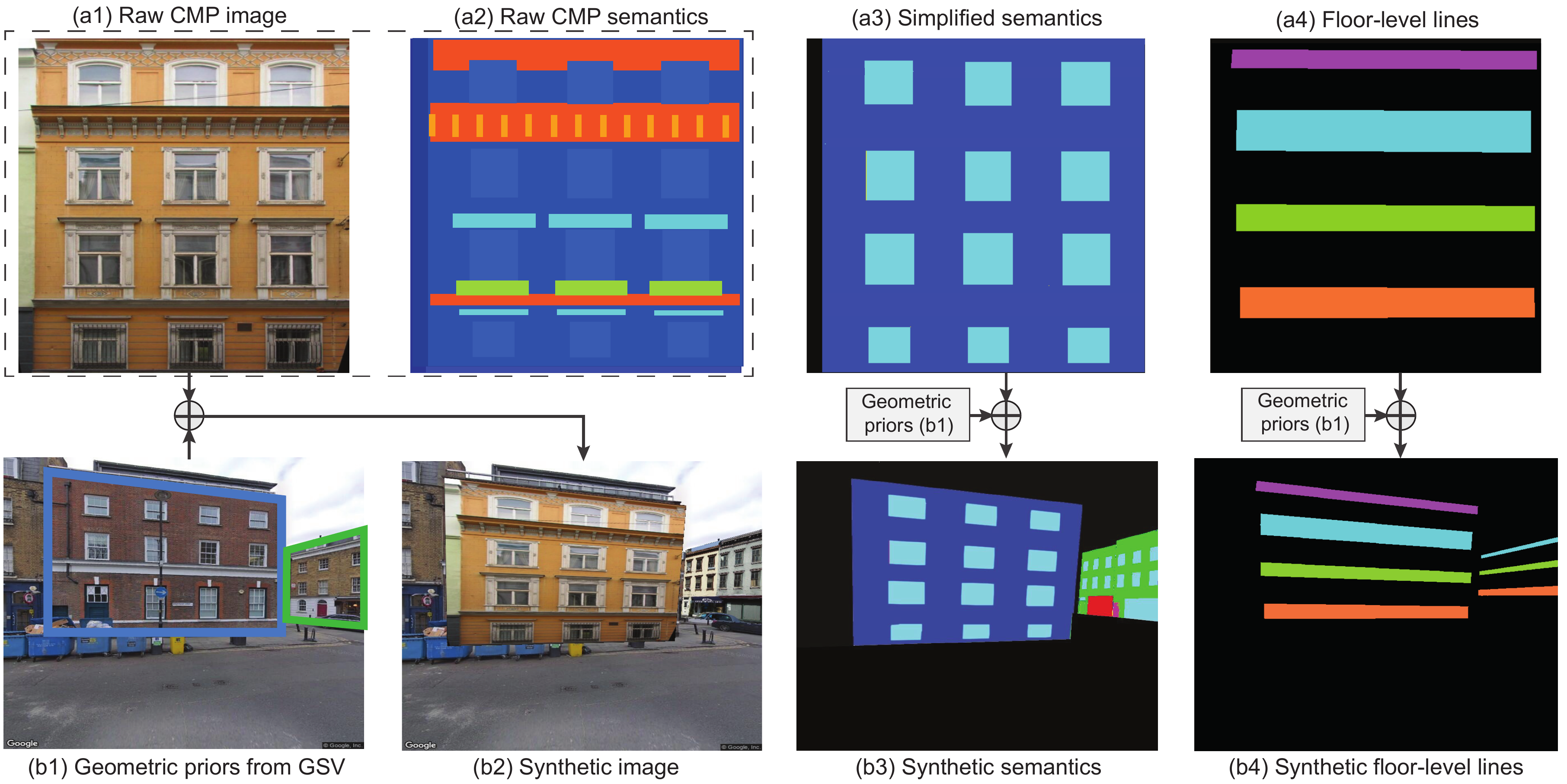}
\vspace{-2mm}
\caption{
Illustration of our data augmentation scheme.
We take the advantages of the rich context of {facades} in the CMP dataset~\cite{Tylecek13} (a1 \& a2) and the perspective-oriented building facades readily extracted from the GSV (Google Street View) images (b1).
From them, we can efficiently obtain simplified semantics (a3) and annotate floor-level lines (a4), and further generate a very large amount of augmented image samples, \ie, an augmented image (b2) with its associated semantic image (b3) and floor-level-lines image (b4), by pairing up different geometric priors with different CMP facades.
}
\label{fig:dataset}
\vspace{-3mm}
\end{figure*}

There are three main parts in our approach.

\begin{enumerate}
\item[(i)]
We compile a set of facade images from the CMP dataset~\cite{Tylecek13}, and employ them to augment building facades in GSV images based on the geometric prior of facade perspective (Sec.~\ref{sec:data_composition}).
By our new data augmentation scheme, we can efficiently generate a large amount of image samples to train our network.
\item[(ii)]
We develop a multi-task learning network$-$\emph{FloorLevel-Net} (Sec.~\ref{ssec:level_feature} to~\ref{ssec:implementation}), to segment the input image into piecewise regions of {facades} and to detect candidate pixels possibly associated with floor-level lines; see the multi-task learning module in Fig.~\ref{fig:overview}.
Particularly, \emph{FloorLevel-Net} encapsulates semantic (\eg,~\emph{windows}, \emph{doors}, \emph{shops}, etc.) and geometric (\eg, \emph{floor orders}, \emph{facade orientation}, etc.) information related to floor-level lines by articulating a fused loss function that is differentiable by means of pixel-wise convolutions.
\item[(iii)]
We infer per-line 5-tuple parameters from the piecewise segmentations in the geometry post-processing stage (Sec.~\ref{ssec:line_params}).
Here, we locate facade regions, group floor-level lines per {facade}, then regress a polyline per floor-level line; see the geometry processing module in Fig.~\ref{fig:overview}.
Further, we refine the polylines based on self-constrained geometric priors, \ie, the {facade} boundary, floor order, and vanishing points.
\end{enumerate}

So far, we are not aware of any work on recognizing floor-level lines in street-view images.
Existing related works focus on recognizing building facades as a whole or explicit objects, such as windows, doors, and balconies.
Hence, we show the effectiveness of our approach through comparisons with ablation techniques (Sec.~\ref{ssec:ablation_analysis}) and DeepFacade, which is a closely-related work on facade parsing (Sec.~\ref{ssec:compare_facade}), and demonstrate the potential of our work to support and enrich urban AR applications (Sec.~\ref{ssec:support_ar}).
\section{Dataset Preparation \& Analysis}
\label{sec:data_composition}

An immediate challenge to recognizing floor-level lines is the lack of properly-annotated street-view images.
On the one hand, existing city-wide datasets such as Cityscapes~\cite{cordts_2016_cityscapes} provide mainly urban scene segmentations, \eg, roads, buildings, vehicles, etc., while lacking details of facades, not to mention floor-level lines.
On the other hand, building facade datasets such as CMP~\cite{Tylecek13} provide detailed semantics of facades, \emph{e.g.}, windows, doors, balcony, etc.
However, the images exhibit mostly rectified views of front-facing facades, which cannot reflect real-world facades with perspective orientations.

\subsection{Data Augmentation}
\label{ssec:data_aug}
To relax the enormous workload for labeling a new dataset, we propose a data augmentation scheme (Fig.~\ref{fig:dataset}) to take the best advantages of the facade semantics in the CMP dataset~\cite{Tylecek13} together with the perspective-oriented building facades extracted from GSV images.

To start, we first analyze the facade images (Fig.~\ref{fig:dataset} (a1)) with their semantics (Fig.~\ref{fig:dataset} (a2)) provided by the CMP dataset.
Here, we make use of the simplified semantics (Fig.~\ref{fig:dataset} (a3)), \ie, \emph{window}, \emph{shop}, and \emph{door}, which offer contextual information for floor-level lines, \emph{i.e.}, windows usually appear on every floor, whereas shops and doors usually appear on the ground floor.
There may not be obvious lines between adjacent floors, yet humans can infer them based on the associated contexts,~\eg, windows.
Thanks to the regular front-facing structures in these inputs, manually labeling floor-level lines (Fig.~\ref{fig:dataset} (a4)) is very fast, compared with labeling on general street-view images that are perspective.
We extract a total of 150 rectified facade images from the CMP dataset.

Next, we employ the GSV API~\cite{gsv_api} to collect street-view photos.
Here, we randomize the camera headings to obtain perspective-oriented facades, and set the pitch parameter in the API to zero, so the camera view directions are always horizontal, \ie, parallel to the ground.
Also, for each GSV photo, we identify nearby planar facades that are feasible for overlaying CMP facade images, \emph{i.e.}, with an aspect ratio similar to those in the CMP dataset.
Then, we manually annotate a quad to mark the region of each feasible planar facades (Fig.~\ref{fig:dataset} (b1)) and label also its orientation 
$\{left, right, front\}$ relative to the camera view direction.
The facade priors serve as reference locations for overlaying facade images and semantics.
For instance, see Fig.~\ref{fig:dataset} (b1), the blue quad marks the facade region of the left building with orientation categorized as \emph{front} (\ie, front facing the camera), whereas the green quad marks the facade region of the right building with \emph{right} orientation.
By doing so, we can derive an affine transformation matrix for each facade, and use it to generate an augmented street-view image (Fig.~\ref{fig:dataset} (b2)) with its associated semantic image (Fig.~\ref{fig:dataset} (b3)) and labeled floor-level-lines image (Fig.~\ref{fig:dataset} (b4)).

To enrich the data sample diversity and to improve the network generalizability, we collect 200 GSV images in Hong Kong and London, with both high-rise and relatively low-rise buildings, respectively.
Together with the facade images from the CMP dataset, we compile a new dataset with 3,000 pairs of augmented street-view images, each associated with facade semantics and floor-level line labels.
We train our \emph{FloorLevel-Net} using the dataset, and compare it with an ablation model trained on the CMP dataset of only rectified facades.
The results show a significant boost by the data augmentation scheme (Sec.~\ref{ssec:ablation_analysis}).

\begin{figure}[t]
\centering
\includegraphics[width=0.99\linewidth]{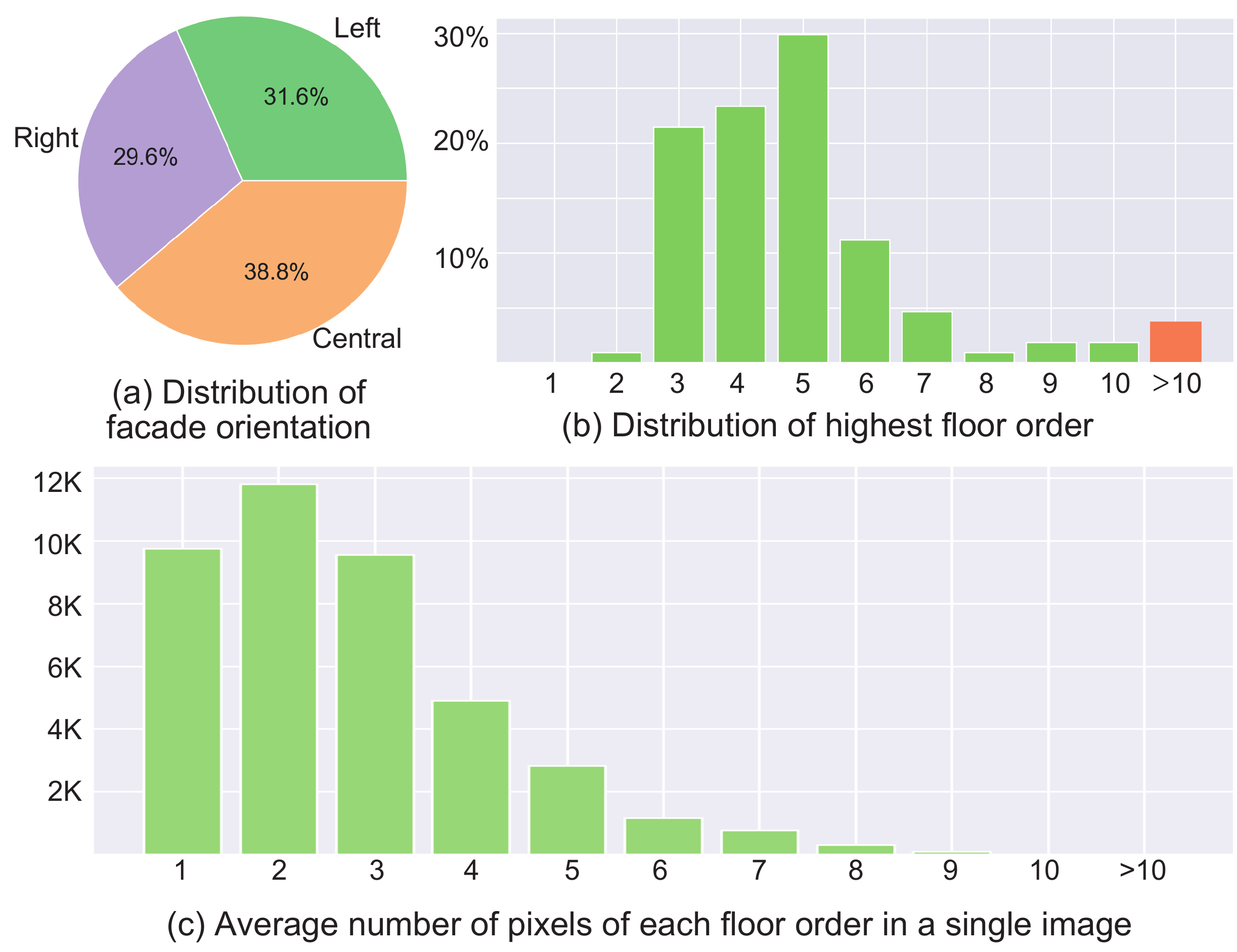}
\vspace*{-7mm}
\caption{
Characteristics of the data samples in our dataset: 
distributions of (a) the facade orientation, (b) the highest floor order, and (c) average number of pixels of each floor order in a single image.}
\label{fig:dataset_stat}
\vspace{-3mm}
\end{figure}

\subsection{Dataset Characterization}
\label{ssec:preliminary}

We conduct a preliminary analysis of our augmented dataset to show its strong generalizability, and later take it to derive feasible components in our network.
By observing the following characteristics of the dataset, some hyper parameters, \eg, maximum floor order, are determined. 

\begin{itemize}

\item
\emph{Facade orientation}:
Fig.~\ref{fig:dataset_stat} (a) reports the distribution of facade orientations.
As mentioned in the previous subsection, three orientations $\{left, right, front\}$ are considered based on the facade orientation relative to the camera.
The three orientations share similar proportions, without biasing towards any of the three.

\item
\emph{Highest floor order}:
Fig.~\ref{fig:dataset_stat} (b) plots the distribution of the highest floor order in the data samples.
Most facades have a floor order of five.
Also, there is no single-floor building, since we omit single-floor buildings when compiling the data.
There are few high-rise buildings ($< 5\%$) with the highest floor order above 10.

\item
{\emph{Average pixel number per floor order}: 
Fig.~\ref{fig:dataset_stat} (c) presents the average number of pixels belonging to each floor order in a single image of size 480$\times$360. Since the street-view images are taken from the ground, low floor orders have more pixels, and the number of pixels decreases as floor order increases. Floor-level lines of orders above 10 are too small to overlay AR images.}

\end{itemize}
In consideration of the distribution of highest floor order and average pixel number per floor order in the augmented dataset, we set the maximum floor order to be 10 in our floor-level line detection module (Sec.~\ref{ssec:level_feature}).

\begin{figure*}[ht]
\centering
\includegraphics[width=0.98\linewidth]{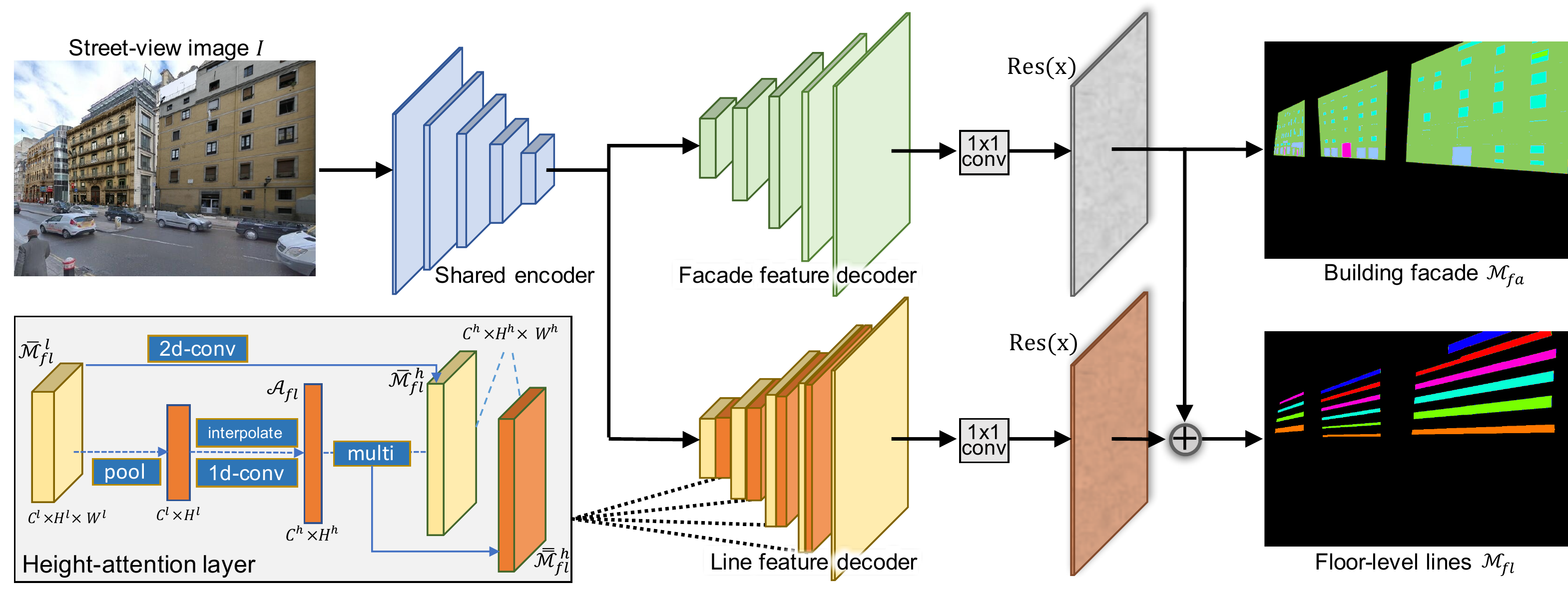}
\vspace{-2mm}
\caption{Overview of our \emph{FloorLevel-Net} framework.
We consume each input street-view image by a multi-task learning network to jointly segment the {facades} and detect floor-level lines.
In the module of our line feature decoder, we incorporate height-attention layers (see bottom left) along with the original convolutional layers, to help enforce the vertical ordering of the floor-level lines, where ``multi'' denotes element-wise multiplication.
}
\label{fig:arch}
\vspace{-2mm}
\end{figure*}

\section{FloorLevel-Net Approach}
\label{sec:floorlevel}

In this section, we first present the framework of \emph{FloorLevel-Net} (see Fig.~\ref{fig:arch}), which is a multi-task learning network (Sec.~\ref{ssec:level_feature}) that fully utilizes the two-stream semantics of {facades} and floor-level lines, with a height-attention mechanism (Sec.~\ref{ssec:HA}) to enforce the vertical ordering of floor-level lines.
Then, we present the implementation details of \emph{FloorLevel-Net} (Sec.~\ref{ssec:implementation}), followed by the add-on geometry post-processing module to generate the final parameters for each floor-level line (Sec.~\ref{ssec:line_params}).

\subsection{Multi-task Learning} 
\label{ssec:level_feature}

Based on the establishment of our augmented dataset, each street-view image $\mathcal{I} \in \mathbb{R}^{W \times H \times 3}$ for training is coupled with two label classes: (i) {facade} semantics $\Psi_{fa}$ and (ii) floor-level distributions $\Psi_{fl}$.
$\Psi_{fa}$ consists of $\{$\emph{window, door, shop, left, right, front, other}$\}$, where the first three labels are the context, the subsequent three indicate the {facade} orientation, and $other$ marks the non-facade pixels.
$\Psi_{fl}$ consists of $\{l_1,\cdots,l_n, other\}$, where $n$ denotes the highest floor order (\ie, 10) as suggested by the data characteristics (Sec.~\ref{ssec:preliminary}), and $other$ marks the non-floor-level-lines pixels.
The two-stream semantics complement each other for the goal:
$\Psi_{fa}$ provides rich context for separating floor-level lines in each {facade}, whereas $\Psi_{fl}$ helps to estimate the order and position of each floor-level line. 
This observation on our training data inspires us to apply a multi-task learning process to benefit each task. 

As illustrated in Fig.~\ref{fig:arch}, \emph{FloorLevel-Net} first employs a shared encoder to learn a feature map, then it applies two-branch decoders to gradually upsample the feature map via deconvolutional layers.
The facade feature decoder in the upper branch outputs segmentation mask $\Bar{\mathcal{M}}_{fa} \in \mathbb{R}^{W \times H}$, 
such that $\Bar{\mathcal{M}}_{fa}(\psi_{fa} | x)$ 
indicates the probability of pixel $x$ having label $\psi_{fa} \in \Psi_{fa}$.
$\Bar{\mathcal{M}}_{fa}$ is computed by
\begin{equation}
\label{eq:bu}
	\Bar{\mathcal{M}}_{fa}(\psi_{fa} | x) = softmax(W_{fa} F(x)),
\end{equation}

\noindent
where $F(x)$ is the feature vector of pixel $x$ and $W_{fa}$ denotes the parameters learned by the network.
We use the loss function $\mathcal{L}_{fa} = \mathcal{C}e(\Bar{\mathcal{M}}_{fa}, \mathcal{M}_{fa})$ to supervise the network training, where $\mathcal{M}_{fa}$ is the ground truth map of {facades} and $\mathcal{C}e(\cdot)$ is a standard softmax cross entropy function.

On the other hand, the line feature decoder in the lower branch of \emph{FloorLevel-Net} outputs floor-level distribution map $\Bar{\mathcal{M}}_{fl} \in \mathbb{R}^{W \times H}$ for floor-level line detection.
Besides the softmax cross entropy $\mathcal{C}e(\Bar{\mathcal{M}}_{fl}, \mathcal{M}_{fl})$, where $\mathcal{M}_{fl}$ denotes the associated ground truth, we also fuse $\mathcal{L}_{fa}$ here in the loss function for supervising the network training:
\begin{equation}
\label{eq:loss}
	\mathcal{L}_{fl} = \mathcal{C}e(\Bar{\mathcal{M}}_{fa}, \mathcal{M}_{fa}) + \mathcal{C}e(\Bar{\mathcal{M}}_{fl}, \mathcal{M}_{fl}).
\end{equation}

Incorporating $\mathcal{C}e(\Bar{\mathcal{M}}_{fa}, \mathcal{M}_{fa})$ in $\mathcal{L}_{fl}$ contributes to the floor-level lines predictions, since facade context such as
windows can help infer the floor levels.
Experimental results also confirm that the fused loss helps to predict more continuous segmentations and more accurate floor orders (Sec.~\ref{ssec:ablation_analysis}).

\begin{figure}[t]
\centering
\includegraphics[width=0.99\linewidth]{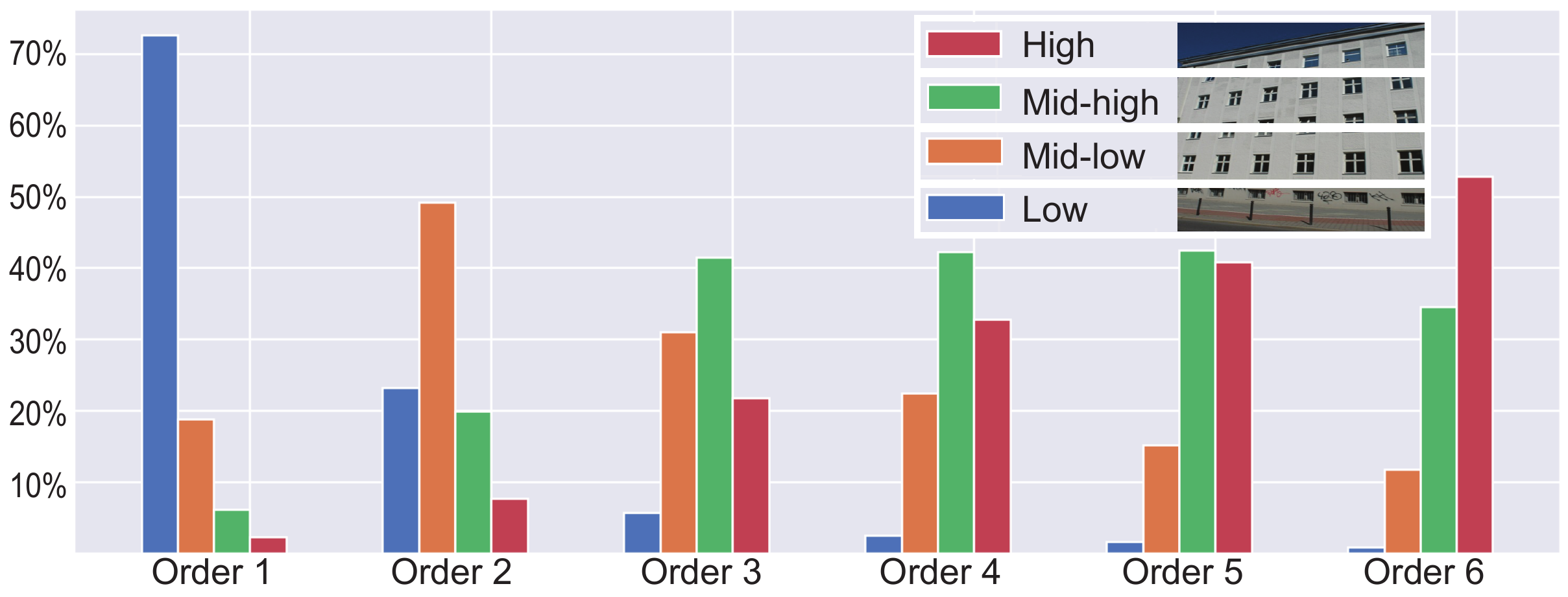}
\vspace*{-7mm}
\caption{
Height-related feature: Pixel distribution of floor-level lines in vertical bounds of the image space.
}
\label{fig:data_ha}
\vspace{-3mm}
\end{figure}

\subsection{Height-Attention Mechanism}
\label{ssec:HA}

Observing that \emph{``cars can't fly in the sky''}, \ie, pixels of \emph{car} are generally below those of the \emph{sky}, Choi et al.~\cite{Choi_2020_HANet} suggest a height-attention mechanism based on the analysis of the CityScape~\cite{cordts_2016_cityscapes} dataset, which indicates that the class distribution significantly depends on a vertical position in urban scene. For instance, a lower part of an image is mainly composed of road, while the upper part, buildings and sky are principal objects.
We hypothesis a similar pattern for pixel vertical distribution of floor-level lines in our case, \ie, pixels of order $i$ floor-level lines are generally below those of order $j$ floor-level lines, given $i < j$.
Here we simplify the analysis by dividing the image space (from bottom to top) into four equal-sized vertical bounds, \ie, \emph{low}, \emph{mid-low}, \emph{mid-high}, and \emph{high}; see the inset in Fig.~\ref{fig:data_ha}.
For each floor-level line in our data samples, we count its pixels in each vertical bound, and compute its distribution ratios in the four bounds.
Then, we sum and normalize the distribution ratios for floor-level lines of same order in the whole data, and obtain the distribution ratios per floor-level line order.
From Fig.~\ref{fig:data_ha}, we can see that pixels of low floor-level lines appear more in the lower vertical bound (\eg, 73\% of floor order 1 pixels are in the \emph{low} bound), whilst pixels of high floor-level lines have a higher chance of locating in the upper vertical bound (\eg, 53\% of floor order 6 pixels are in the \emph{high} bound).

{Next, we analyze the probabilities of floor-level line pixels in the whole image and in each vertical bound.
Table~\ref{table:entropy} presents the probability distributions of the lowest six dominant floors orders, which take over 97\% of all floor orders (see Fig.~\ref{fig:dataset_stat} (c)).
Here $p_{i}$ denotes the probability that an arbitrary pixel is assigned to the $i$-th floor order.
The results further confirm the hypothesis of the pixel vertical distribution.
For example, the probability of floor order 1 $p_{1}$ is 9.75\% in the whole image, and it drops from 27.7\% to 1.8\% in vertical bounds from \emph{low} to \emph{high}.
On the contrary, $p_{6}$ increases from 0.03\% to 3.09\%, which matches with the distribution in Fig.~\ref{fig:data_ha}.
We further measure the uncertainty of pixel distribution probabilities in separate bound by calculating the entropies as $-\sum \mathit{p}_{i}log\mathit{p}_{i}$.
The overall entropy of the entire images is 0.436, and the \emph{low} bound has the smallest entropy of 0.298 due to the dominant probability of floor order 1.
The result indicates that if a pixel falls in the \emph{low} bound, it shall probably be predicted as floor order 1 but not the other floor orders.
}
\begin{table}[]
\centering
\ra{1.3}
\caption{Probability distributions of pixels (in percentage) of the lowest six floor orders in the whole image and in each vertical bound.}
\begin{tabular}{c|cccccc|c|c} \toprule
	\multirow{2}{*}{Given} & \multicolumn{6}{c}{Probabilities of the lowest six floor orders} \vline & \multicolumn{2}{c}{Class} \\ \cline{2-7} 
        & $p_{1}$     & $p_{2}$    & $p_{3}$   & $p_{4}$    & $p_{5}$    & $p_{6}$   & \multicolumn{2}{c}{Entropy} \\ \midrule\midrule
Image    & 9.75      & 11.8      & 9.56     & 4.90     & 2.83     & 1.17    & \multicolumn{2}{c}{0.436}       \\\midrule
Low      & 27.7      & 10.1       & 1.71      & 0.35      & 0.12      & 0.03      &  0.298  & \multirow{4}{*}{\begin{tabular}[c]{@{}c@{}}0.389\\ (avg)\end{tabular}}    \\ 
Mid-low & 6.75      & 20.1      & 8.72      &  2.97     & 1.09      & 0.32      &  0.386    &                      \\
Mid-high & 2.79      & 10.3      & 14.8      &  7.07     & 3.86      & 1.20      &  0.429  &                 \\
High      & 1.80      & 6.66      & 13.0     &  9.20     & 6.22      & 3.09      &  0.442     &     \\ \bottomrule 
\end{tabular}
\label{table:entropy}
\vspace{-4mm}
\end{table}

These observations suggest that vertical position in image space can serve as a good indicator of floor order.
Hence, we are motivated to leverage height-attention layers in the design of \emph{FloorLevel-Net}.
In detail, we include an HA layer between adjacent convolutional layers in the line feature decoder; see Fig.~\ref{fig:arch} (bottom left) for the illustration.

Specifically, the HA layer bridges a lower-level feature map of floor-level lines $\Bar{\mathcal{M}}^{l}_{fl} \in \mathbb{R}^{C_{l} \times W_{l} \times H_{l}}$ and a higher-level feature map $\Bar{\mathcal{M}}^{h}_{fl} \in \mathbb{R}^{C_{h} \times W_{h} \times H_{h}}$ (note: subscripts $l$ and $h$ denotes lower-level and higher-level, respectively) in the following ways:
(i) adopts a width-wise pooling to extract vertical features from $\Bar{\mathcal{M}}^{l}_{fl}$;
(ii) employs a 1-D convolutional layer with a bilinear interpolation to generate attention map $\mathcal{A}_{fl} \in \mathbb{R}^{C_{h} \times H_{h}}$, which shares the same channel and height size as $\Bar{\mathcal{M}}^{h}_{fl}$; and
(iii) generates a refined higher-level feature map $\Bar{\Bar{\mathcal{M}}}_{fl}^{h}$ by an element-wise multiplication of $\mathcal{A}_{fl}$ and $\Bar{\mathcal{M}}^{h}_{fl}$.
In doing so, $\mathcal{A}_{fl}$ enriches the per-channel scaling factors for each individual row of vertical positions, and the refined higher-level feature map $\Bar{\Bar{\mathcal{M}}}_{fl}^{h}$ embodies the height-wise contextual information for locating and also ordering the floor-level lines.

\begin{figure}[t]
\centering
\includegraphics[width=0.985\linewidth]{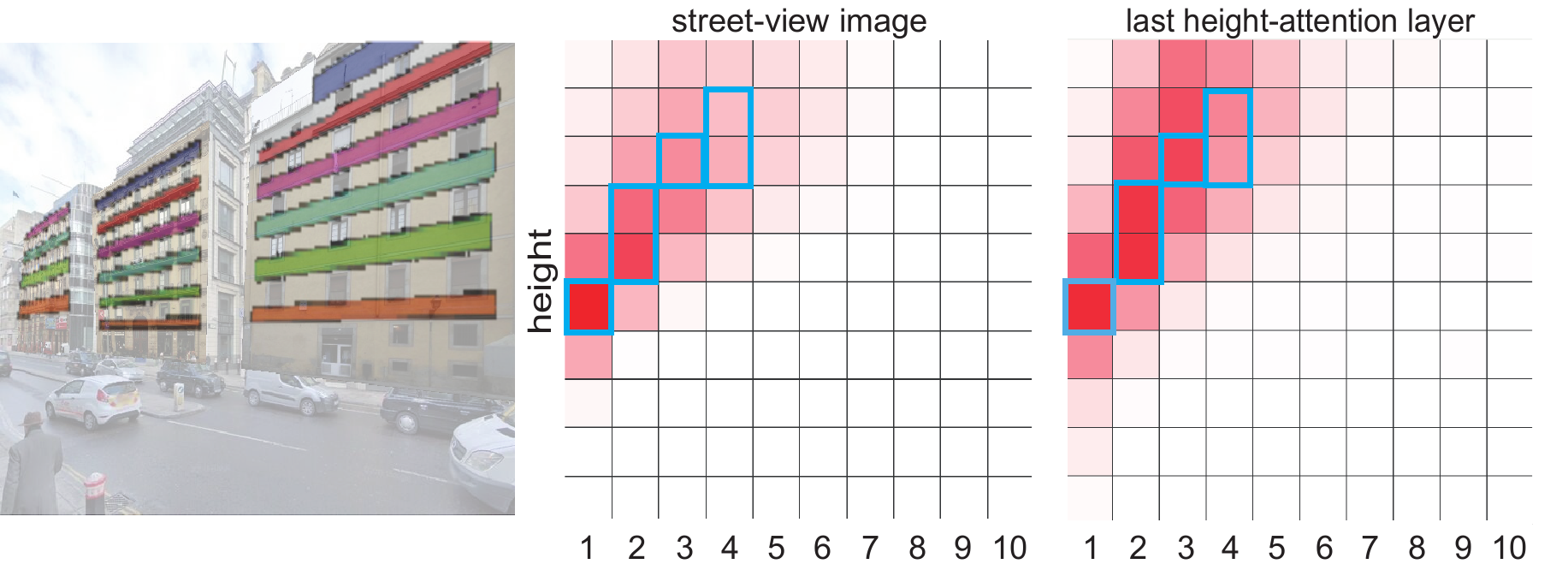}
\vspace{-2mm}
\caption{
\textbf{Left} is an example input image with annotated floor orders. \textbf{Middle} plots the floor-level line pixel distributions in the input image, whereas \textbf{right} plots the attention weights in the last height-attention layer.
}
\label{fig:ha_map}
\vspace{-4mm}
\end{figure}

To study the effectiveness of the HA mechanism, we extract attention weights learned by the last HA layer and compare them with floor-level distributions in street-view images.
Fig.~\ref{fig:ha_map} shows a typical comparison example using the input image in the left.
Here, we plot two heatmaps with floor orders as the horizontal axis and image-space height as the vertical axis.
The heatmaps are constructed in the following way.
Horizontal axes denote floor order from 1 to 10, whereas vertical axes denote vertical position (height) sub-divided into 10 vertical bounds for visualization.
For each cell $(m,n)$, we count the pixels in floor-level $m$ (middle) or extract the attention weight of the $m$-th channel (right), inside the $n$-th vertical bound.
Also, we normalize the intensities (in red) in each map.
Comparing the two heatmaps, we can see that they exhibit very similar patterns.
Particularly, if we mark the most frequent height(s) per floor order (see the blue boxes) in both plots, we can see that the attention weights reveal the heights of the pixels in floor-level lines of different orders.
The comparison together with the ablation analysis presented in Sec.~\ref{ssec:ablation_analysis} demonstrate the effectiveness of the HA mechanism.

\begin{figure*}[ht]
\centering
\includegraphics[width=0.85\linewidth]{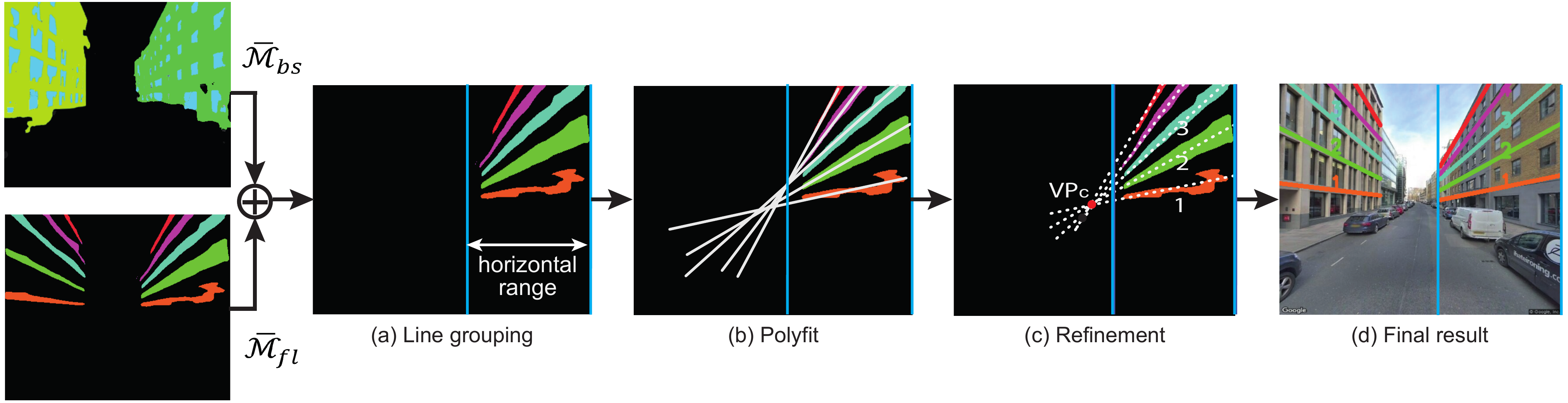}
\vspace{-2mm}
\caption{
Geometry-constrained post-processing for retrieving line parameters: grouping floor-level lines based on {facade} segmentation (a), fitting polylines according to line predictions (b), refining vanishing point using geometry constraints (c), and the final output (d).}
\label{fig:geo_process}
\vspace{-2mm}
\end{figure*}

\subsection{Implementation of FloorLevel-Net}
\label{ssec:implementation}
\emph{FloorLevel-Net} adopts an encoder-decoder structure based on DeepLabV3+~\cite{Chen_2018_Deeplab} with ResNet101~\cite{He_2016_Resnet} as the backbone.
Specifically, we employ four ResNet stages and one atrous spatial pyramid pooling (ASPP) layer to generate the shared features, and use five convolutional layers in each of the decoders for {facade} segmentation and floor-level lines detection.
In the line feature decoder, we arrange one height-attention layer between each pair of adjacent convolutional layer, and append a residual layer at the end to fuse the features from the facade feature decoder.
We use ReLU for all layers, except for the final prediction layers, where softmax is applied.
We implement \emph{FloorLevel-Net} using PyTorch and train it from scratch on a single NVidia GeForce GTX 1080 Ti GPU card.
During the training, we adopt the momentum optimizer with learning rate 1e-3 and a small batch size of four.
Convergence is reached at about 100K iterations.

\subsection{Geometry Post-Processing}
\label{ssec:line_params}
\emph{FloorLevel-Net} predicts two piecewise segmentation masks $\Bar{\mathcal{M}}_{fa}$ and $\Bar{\mathcal{M}}_{fl}$, from which we further extract line parameters by considering various geometric constraints:

\begin{itemize}
\item[(i)]
\textit{Facade constraint}: Floor-level lines are attached to {facades}, so the detected lines are valid only if they lie inside 
a {facade} region.
Since $\Bar{\mathcal{M}}_{fa}$ may contain multiple {facades}, we process floor-level line segmentations and generate only one line with a specific floor order per {facade}.
\item[(ii)]
\textit{Vanishing point (VP) constraint}: Floor-level lines of the same {facade} are intrinsically parallel in 3D, so they should meet at a common VP, which is at a finite location for perspective-oriented {facades}.
\item[(iii)]
\textit{Order constraint}: Assuming that the up direction in input image roughly matches the up direction in real world, orders of floor-level lines for the same {facade} should strictly increase from bottom to top.
\end{itemize}

To enforce these constraints, we formulate a second-stage geometry post-processing to generate floor-level line parameters with the following steps (as illustrated in Fig.~\ref{fig:geo_process}):
\begin{itemize}
\item[(i)]
\textit{Line grouping} (Fig.~\ref{fig:geo_process} (a)):
First, we group relevant labels in $\Bar{\mathcal{M}}_{fa}$ to form a super-segment per {facade}.
Then we group floor-level lines segmentation in $\Bar{\mathcal{M}}_{fl}$, and identify a group of lines per {facade}.

\item[(ii)]
\textit{Polyfit} (Fig.~\ref{fig:geo_process} (b)):
We fit a polyline (in the form of $\beta_{0}^{i} + \beta_{1}^{i}x$) for pixels of each floor-level line (say $\{(x^i_j, y^i_j)\}^n_{j=1}$ for $i$-th line) using a least squares regression:
\begin{equation}
\label{eq:least_square}
	\underset{\beta_{0}^{i}, \beta_{1}^{i}}{\arg\min} \{ \sum_{j=1}^{n}(y^i_{j}-(\beta_{0}^{i} + \beta_{1}^{i}x^i_j))^2 \}.
\end{equation}

Note that on the same {facade}, there could be disjoint predicted regions of the same floor order; we have to merge these regions before the regression.

\item[(iii)]
\textit{Refinement} (Fig.~\ref{fig:geo_process} (c)):
Polylines derived for the same {facade} may not perfectly meet at a common VP, so we regress location $(x_c, y_c)$ for the common $VP_{c}$ and 
refine each polyline as $\beta^{i}(x - x_{c}) + y_{c}$ by taking $VP_{c}$ as an anchor point.
To do so, we formulate a gradient-based optimization to update parameters $\{\beta^{i}\}_{i=1}^m$, $x_{c}$, $y_{c}$ with a global least squares:
\begin{equation}
\label{eq:vp_constrained}
	\underset{\{\beta^{i}\}, x_{c}, y_{c}}{\arg\min} \{ \sum_{i=1}^{m}\sum_{j=1}^{n}(y^{i}_{j}- \beta^{i}(x - x_{c}) - y_{c})^2 \}.
\end{equation}

\item[(iv)]
\textit{Final result} (Fig.~\ref{fig:geo_process} (d)):
Last, we take the horizontal and vertical ranges of each {facade}, to derive the two endpoints ($x_s^i,\, y_s^i$) and ($x_e^i,\, y_e^i$) for each polyline.
Together with the floor order $l^i$, we can then obtain the five-tuple  $\{x^{i}_{s},\, y^{i}_{s},\, x^{i}_{e},\, y^{i}_{e},\, l^{i}\}$ for each floor-level line.

\end{itemize}
\section{Experiment}
\label{sec:experiment}

To the best of our knowledge, this work is the first attempt at recognizing floor-level lines in street-view images.
Hence, there are no benchmark datasets, evaluation metrics, and existing methods for the task.
We fill the gaps by preparing a new dataset (Sec.~\ref{ssec:exp_data}), proposing quantitative metrics for evaluation (Sec.~\ref{ssec:eval}), and comparing with ablated techniques (Sec.~\ref{ssec:ablation_analysis}) and some closely-related techniques (Sec.~\ref{ssec:compare_facade} and Sec.~\ref{ssec:compare_segment}).
Further, we present some AR-related applications to demonstrate the applicability of this work (Sec.~\ref{ssec:support_ar}).

\subsection{Testing Dataset}
\label{ssec:exp_data}

Besides the new training dataset prepared by using our data augmentation scheme (see Sec.~\ref{sec:data_composition}),nwe prepare a new testing dataset for evaluating the effectiveness and generalizability of our approach.
Here, we choose street-view images by considering the followings:
(i) the image should contain at least one (nearby) {facade}, which is not fully occluded;
(ii) the {facades} should contain at least one floor-level line;
(iii) the {facades} can be perspective-oriented in view but cannot be curved in the physical world; and
(iv) the {facades} should contain some semantic elements,~\eg, windows or balconies.
We collect {600} street-view images for the testing dataset from the following four different sources:

\begin{itemize}
\item
{We randomly select 150 street views from the eTrims~\cite{korc-forstner-tr09-etrims} and arcDataset~\cite{xu_2014_architectural}.}
Each image includes only one single {facade}, and the buildings are mostly low-rise.
\item
{We randomly select 150 more images from the LabelMeFacade~\cite{Froehlich-Rodner-Denzler-ICPR2010}, TSG-20~\cite{tsg20} and ZuBuD+~\cite{shao_2003_zubud} datasets.}
These images have more high-rise buildings, and some have multiple {facades}.
\item
We randomly download {200} GSV images in London and Glasgow (UK), which feature typical European-style buildings, similar to those in the CMP dataset.
\item
Lastly, we randomly download another 100 GSV images in Hong Kong (HK).
These images feature high-rise buildings, and many {facades} are partially occluded by billboards, advertisements, etc.
\end{itemize}

\begin{figure*}[t]
\centering
\includegraphics[width=0.99\linewidth]{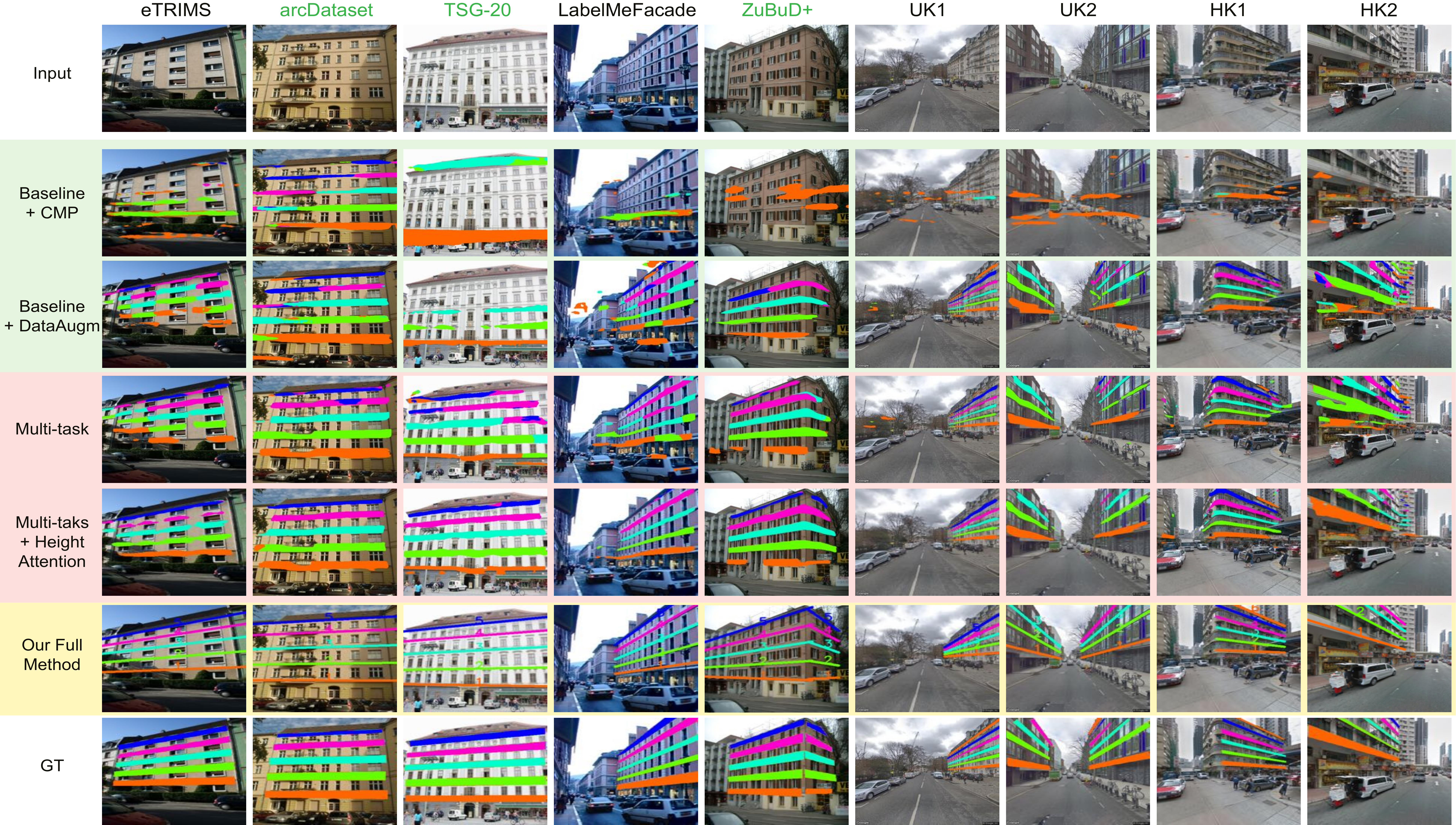}
\vspace*{-2.1mm}
\caption{
Qualitative comparison results in the ablation analysis on the unseen testing dataset.
The results illustrate the effects of (i) our data augmentation scheme by comparing DeeplabV3+ models trained on CMP and on our training data (green background), (ii) multi-task learning with and without the height-attention mechanism (red background), and (iii) our full method further with geometry post-processing (yellow background), vs. the ground truths (GT).}
\label{fig:ablation}
\vspace*{-1mm}
\end{figure*}

On each test image, we manually label the position and order of each floor-level line on the {facades} in the form of a quadrilateral region,~\eg, see the ``GT'' row in Fig.~\ref{fig:ablation}.
{It takes around 2 minutes to label one image and around 20 hours in total for all the 600 images.}
The datasets are available on the project website:
\url{https://wumengyangok.github.io/Project/FloorLevelNet/}.

\subsection{Evaluation Metrics}
\label{ssec:eval}

As a two-stage method, our results include (i) predicted pixel-wise segmentations from \emph{FloorLevel-Net}, and (ii) regressed endpoints $(x^i_s, y^i_s)$ \& $(x^i_e, y^i_e)$ and order $l^i$ of each floor-level line from the geometry post-processing.
We employ the following metrics to evaluate the two results separately:
\begin{itemize}
\item
{\em Pixel-wise accuracy\/}.
For each street-view image, we denote the set of pixels in ground-truth and in predicted line regions as $\mathcal{P}_{gt}$ and $\mathcal{P}_{pred}$, respectively.
We employ the $F_1$ score for quantitative comparisons, which can be measured upon 
the number of true positives (TP = $|\mathcal{P}_{gt} \cap \mathcal{P}_{pred}|$), 
false positives (FP = $|\mathcal{P}_{pred} - \mathcal{P}_{gt} \cap \mathcal{P}_{pred}|$),
and false negatives (FN = $|\mathcal{P}_{gt} - \mathcal{P}_{gt} \cap \mathcal{P}_{pred}|$):
\begin{equation}
\label{eq:pixel_wise_ac}
F_1 = (2TP)/(2TP + FP + FN).
\end{equation}
\item
{\em Line-wise accuracy\/}.
We join the endpoints of each floor-level line $L_i$ to form a three-pixel wide straight line, yielding a bag of pixels $\mathcal{P}_i$.
Then, we can compute the confidence of $L_i$ as being correctly recognized as
\begin{equation}
\label{eq:con_interval}
CI(L_i) = \frac{\sum_{p\in \mathcal{P}_i}{\mathbb{1}(\mathcal{M}_{fl}(p) = l^i)}}{|\mathcal{P}_i|},
\end{equation}

\noindent
where $\mathbb{1}(\cdot)$ is an indicator function for counting the number of pixels in $P_i$ that appears on the ground truth image $\mathcal{M}_{fl}$ with label $l^i$.
If $CI(L_i) > 0.5$, we regard the line prediction as a true positive.
For each street-view image, we again count the number of floor-level lines as being correctly detected (TP), missed (FN), and wrongly recognized (FP), then employ the $F_1$ score (like Eq.~\eqref{eq:pixel_wise_ac}) to compute the line-wise accuracy.

\end{itemize}

\subsection{Ablation Analysis}
\label{ssec:ablation_analysis}

We carry out an ablation analysis to evaluate the major components in our approach, including (i) the data augmentation scheme, (ii) the multi-task learning, (iii) the height-attention mechanism in \emph{FloorLevel-Net}, and (iv) the geometry-post-processing.
In the analysis, we employ the testing dataset, in which the images do not appear in the training data.

\vspace{1mm}
\noindent
\textbf{Ablated techniques}.
To evaluate components in \emph{FloorLevel-Net}, we take a stepwise testing strategy by adding the components one by one to the baseline DeepLabV3+~\cite{Chen_2018_Deeplab} model until we have our full architecture.
The difference between consecutive tests indicates the performance enhancement made by the added component.
To start, we evaluate our data augmentation scheme by the following pair of experiments:
\begin{itemize}
\item
\emph{Baseline+CMP}: We train a DeepLabV3+ model (baseline) using the facade dataset CMP~\cite{Tylecek13} with rectified facades (Fig.~\ref{fig:dataset} (a1)) and our labels of floor-level lines (Fig.~\ref{fig:dataset} (a4)).
\item 
\emph{Baseline+DataAugm}: We train another DeepLabV3+ model on the training dataset of 3,000 pairs of augmented street-view images (Fig.~\ref{fig:dataset} (b2)) and labels of floor-level lines (Fig.~\ref{fig:dataset} (b4)), which are produced by our data augmentation scheme, as described in Sec.~\ref{ssec:data_aug}.
\end{itemize}

Next, we evaluate the multi-task learning framework and height-attention mechanism in \emph{FloorLevel-Net} by adding these components to \emph{Baseline+DataAugm}, as follows: 
\begin{itemize}
\item 
\emph{Multi-task}: This refers to the multi-task learning network that we build to simultaneously predict {facades} and floor-level lines, as described in Sec.~\ref{ssec:level_feature}. 
\item 
\emph{Multi-task+HA}: We further incorporate a set of height-attention layers, as described in Sec.~\ref{ssec:HA}, to form the full architecture of \emph{FloorLevel-Net}. 
\end{itemize}

Last, we arrive at \emph{Our Full Method} by including the second-stage geometry-post-processing to regress floor-level lines.

\vspace{1mm}
\noindent
\textbf{Qualitative analysis}.
Fig.~\ref{fig:ablation} shows the qualitative comparison results.
{Here, we select one or two street-view images from each constituting component in our testing dataset: eTrims (1st column), arcDataset (2nd column), TSG-20 (3rd column), LabelMeFacade (4th column), ZuBuD+ (5th column), GSVs in UK (6th \& 7th columns), and GSVs in Hong Kong (8th \& 9th columns).}
The top row shows the input images, the bottom row shows the ground-truth images, whereas each of the other rows from the top show the prediction results obtained by the ablated methods,~\ie, \emph{Baseline+CMP}, \emph{Baseline+DataAugm}, \emph{Multi-task}, \emph{Multi-task+HA}, and \emph{Our Full Method}.

Overall, \emph{Baseline+DataAugm} generates better predictions than \emph{Baseline+CMP} (comparing 2nd \& 3rd rows in Fig.~\ref{fig:ablation}), especially for street views with multiple {facades} and for {facades} that are perspective-oriented,~\eg, the GSV images in UK \& Hong Kong.
The results indicate the effectiveness of our data augmentation scheme in generating augmented street-view images for network training.

Next, comparing 3rd \& 4th rows Fig.~\ref{fig:ablation}, we can see that \emph{Multi-task} is able to produce more continuous and distinctive segmentations of floor-level lines than \emph{Baseline+DataAugm}.
This shows that exploring the rich semantics in {facades} indeed facilitates the recognition of floor-level lines.
Further, we can see that the full architecture of \emph{FloorLevel-Net} with the height-attention mechanism (5th row in Fig.~\ref{fig:ablation}) can produce even better results, especially in the prediction of floor orders.
Particularly, the full architecture of \emph{FloorLevel-Net} successfully predicts the floor orders, even for challenging street-view images shown in the 4th to 9th columns, whilst the method fails in the absence of the height-attention mechanism.
Better predictions help simplify the line generation and refinement in the subsequent geometry post-processing stage.
Benefiting from the fine predictions by \emph{FloorLevel-Net}, the post-processing stage (6th row in Fig.~\ref{fig:ablation}) is able to deduce satisfactory floor-level line parameters, subject to the various geometric constraints we discussed earlier.

\begin{table*}[h]
\scriptsize
\centering
\caption{{Quantitative comparison with the ablated techniques in terms of pixel-wise accuracy on network prediction results, and line-wise accuracy on geometry post-processing results.
Results colored in blue are associated with our full method.}}
\vspace*{-1mm}
\begin{tabular}{c|l||c|c||c|c||c|c}\hline
&                 &         \multicolumn{2}{c||}{Lower floors} &  \multicolumn{2}{c||}{Upper floors} & \multicolumn{2}{c}{Overall} \\\hline
&                 & Pixel-wise             & {Line-wise}  & Pixel-wise  & {Line-wise}   & Pixel-wise          & {Line-wise} \\
&                 & Accuracy             & {Accuracy}  & Accuracy  & {Accuracy}   & Accuracy          & {Accuracy}    \\\hline
\multirow{4}{*}{\shortstack{Facades \& \\ buildings}}  
										  & Baseline + CMP    & 0.374                 & 0.725      & 0.136     & 0.676     & 0.289   & 0.702  \\
                                          & Baseline + DataAugm    & 0.599                   & 0.887      & 0.479      & 0.828     & 0.560   & 0.859 \\
                                          & Multi-task      & 0.631                   & 0.912      & 0.498      & 0.845     & 0.586   & 0.880  \\
                                          &  {Multi-task + HA}         & \textbf{0.676}              & \textcolor{blue}{\textbf{0.926}}  & \textbf{0.586}      & \textcolor{blue}{\textbf{0.860}}     & \textbf{0.644}   & \textcolor{blue}{\textbf{0.894}}  \\\hline
\multirow{4}{*}{\shortstack{GSV in \\ UK \& HK}}
                                          & Baseline CMP    & 0.152                 & 0.592      & 0.114     & 0.486     & 0.133   & 0.524  \\
                                          & Baseline + DataAugm    & 0.530                  & 0.781      & 0.449      & 0.708     & 0.505   & 0.742  \\
                                          & Multi-task     & 0.555         & 0.819      & 0.491      & 0.734     & 0.535   & 0.774  \\
                                          & {Multi-task + HA}         & \textbf{0.636}          & \textcolor{blue}{\textbf{0.854}}      & \textbf{0.531}      & \textcolor{blue}{\textbf{0.749}}     & \textbf{0.605}  & \textcolor{blue}{\textbf{0.798}} \\ \hline
\end{tabular}
\label{table:ablation}
\vspace{-2mm}
\end{table*}

\vspace{1mm}
\noindent
\textbf{Quantitative analysis}.
Further, we conduct a quantitative evaluation for each ablated technique.
Here, we categorize the testing dataset into two groups: images from datasets focusing on buildings or facades (eTrims, LabelmeFacade, etc.) are relatively easier for recognizing floor-level lines as the images are prepared for facade parsing or building classification task with regular facade elements, whereas GSVs in UK \& HK with complex environments and dynamic attributes are harder for recognition.
We would also like to evaluate the performance of our approach for different floor-level lines, as many AR applications need only floor information close to the camera.
To do so, we categorize floors 1-3 as lower floors since most of these floor pixels are in the low bound of image space, and the remaining floors as upper floors.

\begin{figure}[t]
\centering
\includegraphics[width=1.\linewidth]{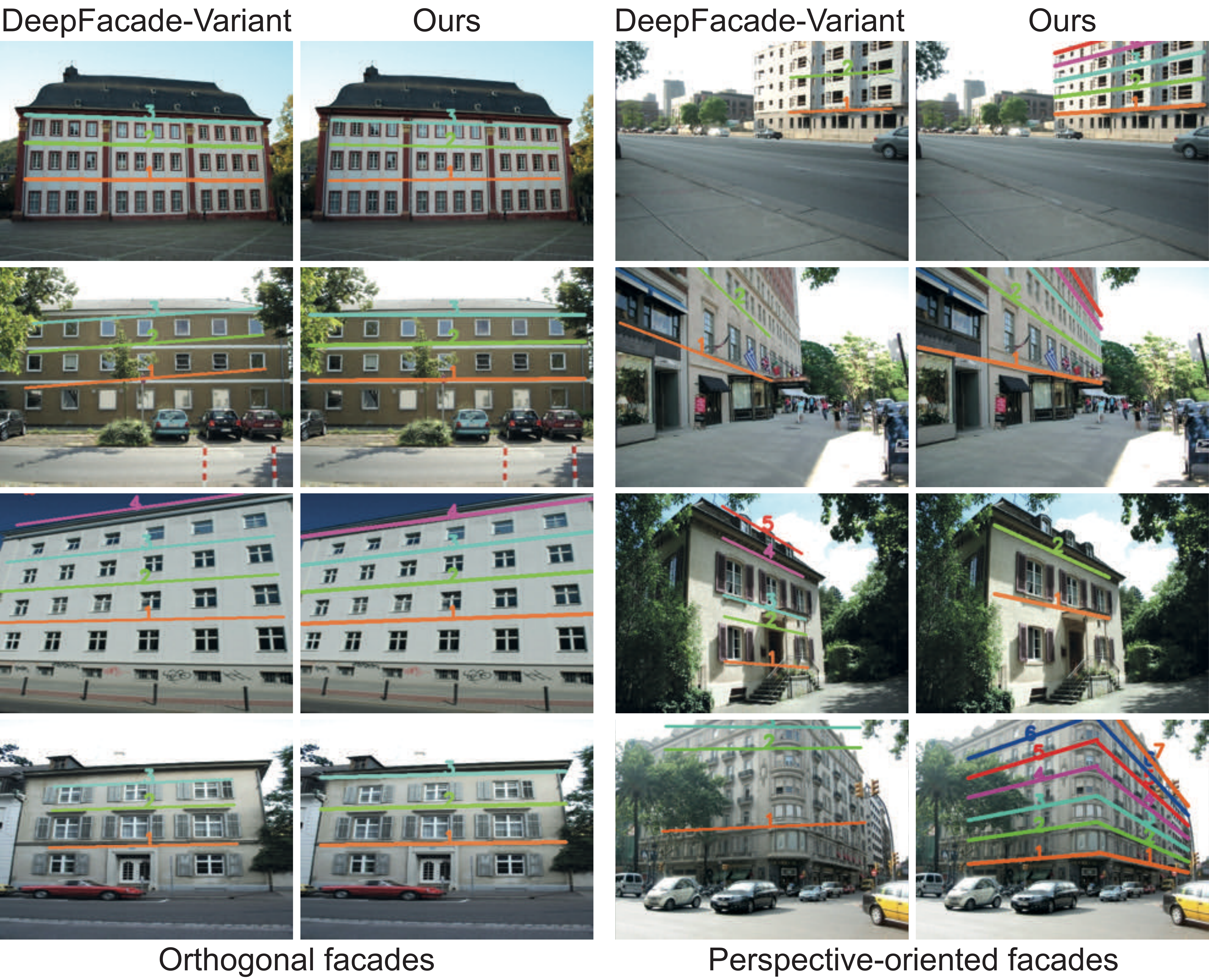}
\vspace{-5mm}
\caption{Comparison between our approach and DeepFacade-Variant on orthogonal (left two columns) and perspective (right two columns) {facades}. Both methods produce satisfactory results for {facades} that are orthogonal to camera view, whilst our approach is more robust to {facades} in perspective view, and multiple facades in one image (last row).
}
\label{fig:cmp}
\vspace{-2mm}
\end{figure}

\begin{figure*}[t]
\centering
\includegraphics[width=0.99\linewidth]{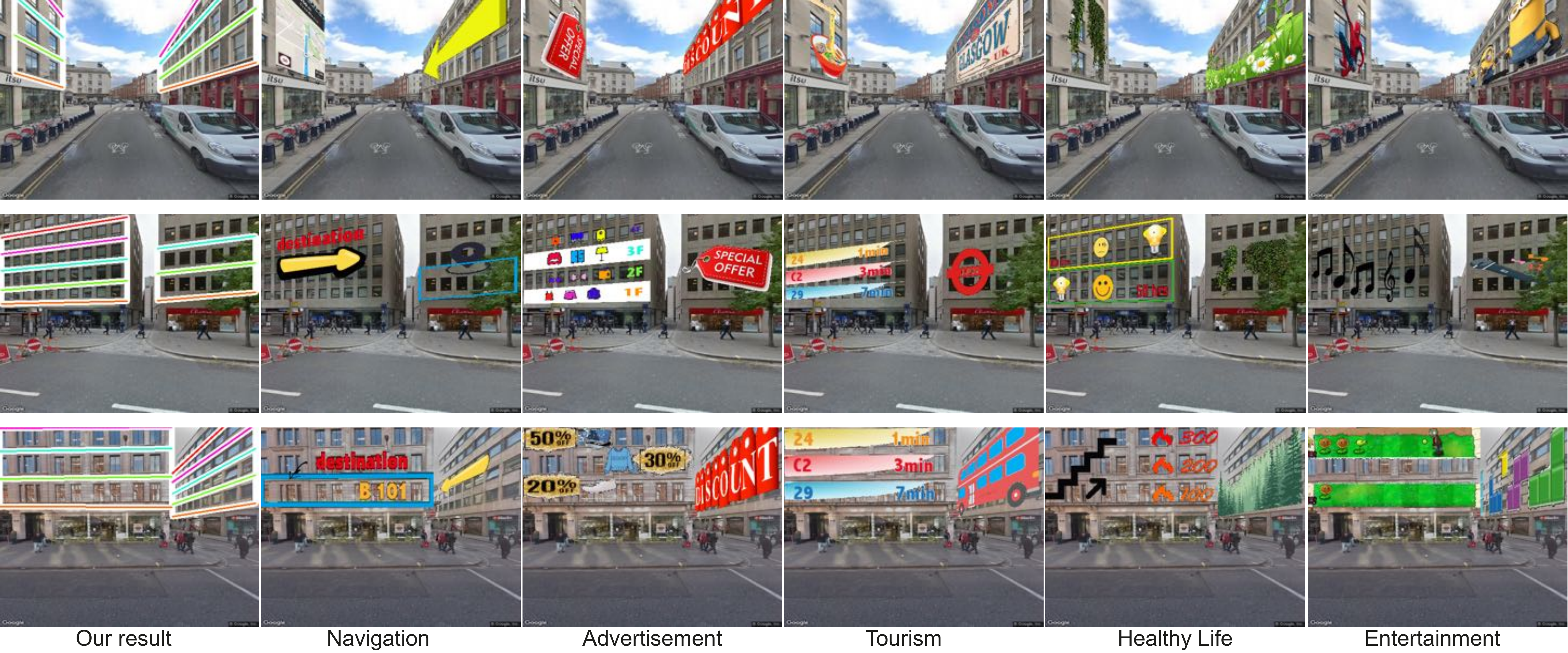}
\vspace{-3mm}
\caption{The potential of our approach to support and enrich various AR scenarios,~\eg, navigation, advertisement, etc.}
\label{fig:ar}
\vspace{-2mm}
\end{figure*}

We measure the pixel-wise accuracy on the ablation network outputs, and line-wise accuracy on  the line parameters after geometry post-processing.
The quantitative results are presented in Table~\ref{table:ablation}, in which our full method, \emph{Multi-task + HA} with geometry post-processing, is marked in blue. 
Based on the results, we can derive some interesting observations.
First, the overall accuracy drops when scene complexity increases from simple street views in facade datasets to GSVs in UK and HK.
There are more {facades} and higher floors in GSVs, especially for those in HK, resulting in less accurate predictions.
Second, predictions for lower floors have better accuracy than those for higher floor.
This is probably because of relatively weaker supervisions for upper floor-level lines, as upper floor-level lines are typically further away from the camera than the lower ones and there are relatively less contextual information.
Third, the data augmentation scheme, multi-task learning, and height-attention layers gradually contribute to the performance enhancement, yet there are still much room for improvement in terms of pixel-wise accuracy. 
The result indicates that floor-level line recognition is more challenging compared with general urban scene segmentation tasks. 
Nevertheless, our predictions provide enough information for the geometry post-processing, which can correctly recognize and locate most floor-level lines in terms of line-wise accuracy.

\subsection{Comparison with a Variant of DeepFacade}
\label{ssec:compare_facade}

As there are no previous works on locating floor-level lines, we consider those on facade parsing,~\ie, segmenting a building facade into windows, doors, etc., which can then serve as references for locating floor-level lines, as competitors.
Here, we compare with DeepFacade~\cite{liu_2017_deep_facade}, which is a state-of-the-art deep-learning-based method for facade parsing.
In detail, we extract floor-level lines from their results as follows:
(i) use RANSAC~\cite{chaudhury_2014_auto} to rectify a street-view image and record the associated homographic matrix;
(ii) apply DeepFacade~\cite{liu_2017_deep_facade} on the rectified image to produce a facade segmentation map, and extract only the window, door, and shop regions as positive regions in the map;
(iii) mark a floor-level line in-between each pair of adjacent (piecewise) positive regions;
and
(iv) apply the inverse of the homographic matrix to revert the original positions of the detected lines to obtain the final results.
From now, we refer the above steps as \emph{DeepFacade-Variant}.

This experiment compares \emph{Our Full Method} with \emph{DeepFacade-Variant}.
However, we find that \emph{DeepFacade-Variant} can hardly produce satisfactory results for GSVs in UK and HK (see supplemental material for examples) for two reasons:
(i) RANSAC rectification is constrained to street views with only a single {facade}, but street views in UK and HK often contain multiple {facades}, leading to multiple vanishing points detected and unstable rectification results;
(ii) buildings in modern cities may contain facades without obvious line segments, thereby making it hard to deduce the vanishing points and perform the rectification.
{Hence, we report the results only on 300 testing images from the eTrims, arcDataset, LabelMeFacade, TSG-20, and ZuBuD+ datasets, which contain only one building facade in most of the images.}

Fig.~\ref{fig:cmp} shows typical results produced by our approach and \emph{DeepFacade-Variant}.
From the results shown on the left (1st \& 2nd columns), we can see that both methods produce good results for {facades} that are orthogonal to camera view.
Yet, \emph{DeepFacade-Variant} is prone to fail when the improper homographic matrix inferred by RANSAC (see row 2 for an example), whilst our approach overcomes the limitation with the refinement made by geometry constraints.
More failure cases for \emph{DeepFacade-Variant} are presented in 3rd \& 4th columns, where the {facades} are in perspective view.
Input image of row 1 is not rectified properly, hence the interpolated lines point to wrong directions.
In rows 2, windows on up floors have little saliency due to distortion, thus the results miss floor-level lines on high levels.
The building in row 3 has partial irregular facade textures, such as lintel and stairs, and the method wrongly recognizes adjacent floors.
In the last row, there are two {facades}, whilst the rectification cannot be managed by one single homographic matrix.
In all these cases, our approach produces good predictions.

\begin{table}[t]
\centering
\caption{{Quantitative comparison between our approach and \emph{DeepFacade-Variant} in terms of accuracy and time efficiency.}}
\resizebox{\columnwidth}{!}{
\begin{tabular}{l|c|cccc} \hline
 & Line-wise & Recti- & Network & Post- & \multirow{2}{*}{SUM}  \\ 
 & accuracy & fication & inference & processing & \\ \hline
DeepFacade-Variant & 0.721 & 1.08s & 0.06s & 0.04s & 1.18s \\
Ours & \textbf{0.876} & - & 0.05s & 0.14s & \textbf{0.19s} \\ \hline
\end{tabular}
}
\label{table:deepfacade}
\end{table}

Table~\ref{table:deepfacade} quantitatively compares the results of \emph{DeepFacade-Variant} and \emph{Our Full Method}.
From the line-wise accuracies, we can see an obvious advancement of our approach over the competitor.
Also, our approach outperforms \emph{DeepFacade-Variant} in terms of execution time, in which the total running time is only 0.19 seconds vs. 1.18 seconds.
In \emph{DeepFacade-Variant}, a major bottleneck is the heavy time cost of RANSAC rectification, whilst our approach directly predicts pixelwise segmentation without requiring rectification.
{
\subsection{Comparison with Semantic Segmentation}
\label{ssec:compare_segment}
As the pixel-wise floor segmentation can be formulated as an image segmentation task, we also present comparison with semantic segmentation methods on the pixel accuracy of the intermediate results. Here we choose PSPNet~\cite{zhao_2017_pspnet} and OCR~\cite{yuan_2020_object} as baselines, since PSPNet is widely used as baseline for semantic segmentation task, and OCR is one of the state-of-the-art methods with highest pixel accuracy in some urban semantic segmentation datasets. For each method, we train the model with both CMP dataset and our augmented dataset, and test with 600 real images following the same settings as the ablation study in Sec.~\ref{ssec:ablation_analysis}.

Table~\ref{table:cmp_segment} presents quantitative results in terms of pixel-wise accuracy. Similar to the results of ablation analysis, the low overall pixel accuracy by the comparison methods trained on CMP dataset indicates the difficulty of inferring floor-level lines, especially without proper dataset. Our data augmentation significantly improves the prediction accuracy to an acceptable level for both segmentation methods, showing well generalizability of the baseline network architectures. Moreover, our methods with multi-task learning and height-attention mechanism achieves the best performance, for both the facades \& buildings datasets and GSVs in UK \& HK.
}

\begin{table}[]
\centering
\caption{{Quantitative comparison between our method and image segmentation baselines in terms of pixel-wise accuracy.}
}
\begin{tabular}{c|l|c|c|c}
                          Datasets                                               & Segmentation methods     & Lower & Upper & Overall \\\hline
\multirow{5}{*}{\begin{tabular}[c]{@{}c@{}}Facades \& \\ buildings\end{tabular}} & PSPNet + CMP      & 0.258 & 0.059 & 0.197   \\
                                                                                 & PSPNet + DataAugm & 0.649 & 0.528 & 0.609   \\
                                                                                 & OCR + CMP         & 0.379 & 0.254 & 0.330        \\
                                                                                 & OCR + DataAugm    & 0.610 & 0.492 & 0.570        \\
                                                                                 & Ours              & \textbf{0.676} & \textbf{0.586} & \textbf{0.644}   \\\hline
\multirow{5}{*}{\begin{tabular}[c]{@{}c@{}}GSV in \\ UK \& HK\end{tabular}}      & PSPNet + CMP      & 0.081 & 0.001 & 0.067   \\
                                                                                 & PSPNet + DataAugm & 0.369 & 0.205 & 0.321   \\
                                                                                 & OCR + CMP         & 0.083 & 0.016 & 0.062        \\
                                                                                 & OCR + DataAugm    & 0.626 & 0.474 & 0.581        \\
                                                                                 & Ours              & \textbf{0.636} & \textbf{0.531} & \textbf{0.605}  
\end{tabular}
\label{table:cmp_segment}
\vspace{-2mm}
\end{table}

\subsection{Applicability}
\label{ssec:support_ar}

Fig.~\ref{fig:ar} shows some example usage scenarios that demonstrates the potential of our method in enriching AR-related applications.
Enabled by the recognized floor-level lines, we can augment real-world scenes with virtual contents in a floor-aware manner.
First, from the floor-level line results, we can identify the bound of each {facade},~\ie, the lowest and highest floor-level lines (and their locations), and overlay virtual contents that cover the whole {facade}, while skipping the real contents on the ground level (Fig.~\ref{fig:ar} (top)).
By this means, we may deliver AR contents, without obscuring necessary real contents around the user on the street.

More importantly, with the floor-level lines, we can place context-aware information on specific floors.
For example, in the bottom-row result on the 2nd column of Fig.~\ref{fig:ar}, we specifically mark the location of the target room for convenient navigation; and in the middle-row result on the 3rd column of Fig.~\ref{fig:ar}, we put shopping directory information over the corresponding floors to aid users to find the user's target items in the department store.
Besides, we may mark specific apartments in a building that are for sales or for rent to aid the property agents.
On the other hand, we may also employ our approach to present floor-related data (\eg, floor size, occupancy information, etc.) on buildings and extend it to estimate the number of floor levels and infer the building heights, which could be helpful for land usage analysis and urban planning~\cite{shen_2017_streetvizor, wu_2018_city-wide,zeng_2018_vitalvizor}.
\begin{figure}[!t]
\centering
\includegraphics[width=0.97\linewidth]{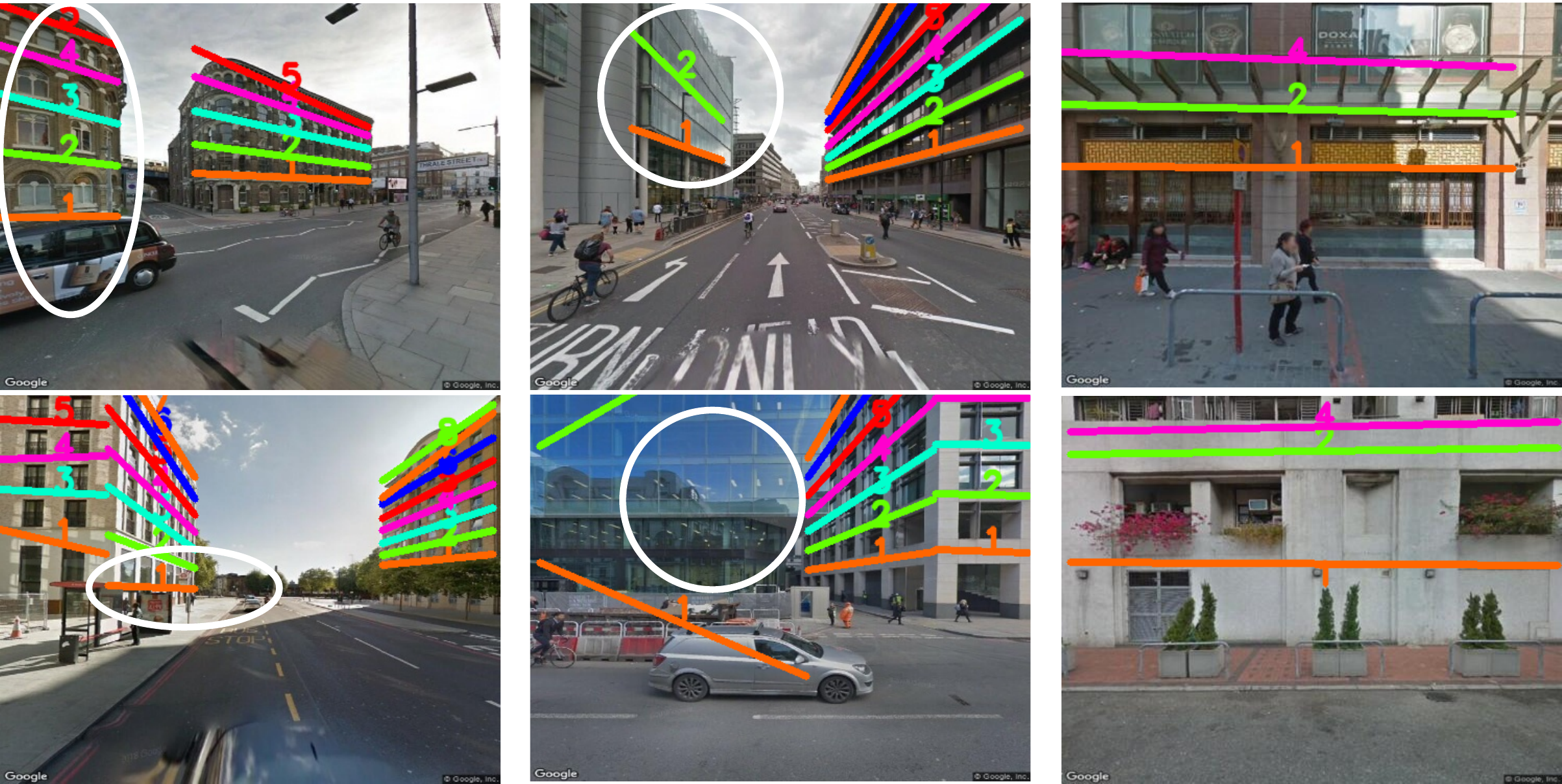}
\vspace{-2mm}
\caption{Limitations of our current approach: obstacle interference (left), glass reflection (middle), and {facades} too close to camera (right).
}
\label{fig:fail}
\vspace{-2mm}
\end{figure}

\section{Conclusion, Discussion, and Future Work}
\label{sec:conclusion}

We presented a new approach for recognizing floor-level lines in street-view images.
As a first attempt to this challenging task, we contribute to
(i) devising a new data augmentation scheme that leverages an existing facade dataset to efficiently generate data samples for network training; and
(ii) formulating \emph{FloorLevel-Net}, a new multi-task learning framework that associates explicit features of {facades} and implicit floor-level lines and incorporates a height-attention mechanism to enforce a vertical ordering of floor-level lines.
The pixelwise semantic segmentations by \emph{FloorLevel-Net} are further refined through a geometry post-processing module for plausible and consistent reconstruction of floor-level lines.
Quantitative and qualitative comparisons with existing methods demonstrate the effectiveness of our proposed method.
Further, we demonstrate the potential of our approach in supporting various AR scenarios.

{
\vspace{1mm}
\noindent
\emph{Discussion.}
This work targets at inferring floor-level lines in street-view images.
We proposed the data augmentation scheme and \emph{FloorLevel-Net} to overcome the challenges of being lack of \emph{training dataset} and \emph{suitable network architecture}.
Though the methods are dedicated to the specific task, we believe the ideas of augmenting dataset and tuning network based on data characteristics can be feasibly extended to some other applications, where labeling is expensive and data have certain patterns. For instance, in autonomous driving, a key requirement is to detect driving lanes and pedestrians simultaneously. In such scenarios, we can possibly improve the network performance by carefully modeling the relationship between driving lane and pedestrian.
}

\vspace{1mm}
\noindent
\emph{Limitations.}
First, our method could be disturbed by obstacles at critical regions,~\eg, a cab that occludes the ground floor of a building (Fig.~\ref{fig:fail} (left)) may cause a wrongly-recognized ground floor.
Second, our method cannot handle {facades} that are reflective (Fig.~\ref{fig:fail} (middle)).
Third, when the camera is too close to the {facades} (Fig.~\ref{fig:fail} (right)), our method cannot get a wide view of the {facades}, so it may wrongly recognize the upper and lower bounds of a floor-level line region as two successive floor-lever lines.

\vspace{1mm}
\noindent
\emph{Future works.}
We plan to focus on improving the prediction precision and efficiency of our approach.
First, a possible direction is to work on short video clips instead of on single images, so that we can obtain temporal information,~\eg, structure from motion (SfM) features, to improve the prediction precision.
Second, we plan to develop strategies to estimate the distance from camera to {facades} to address the last limitation mentioned above.
Third, our current pipeline takes around 0.19 seconds (see Table~\ref{table:deepfacade}), which is not slow compared with DeepFacade-Variant but insufficient to run in real-time.
At present, the main bottleneck is the post-processing module, which can be improved through multiprocessing implementations, since we can simultaneously process line grouping and polyfit.
Also, we plan to explore lightweight networks to further improve the pipeline efficiency.

\section*{Acknowledgment}
We thank the street-view images from the Google Street View service.
This work is supported partially by the Research Grants Council of the Hong Kong Special Administrative Region (Project no. CUHK 14206320) and Guangdong Basic and Applied Basic Research Foundation (2021A1515011700).

\bibliographystyle{abbrv}
\bibliography{main}

\vspace*{-5mm}
\begin{IEEEbiography}[{\includegraphics[width=1in,height=1.25in,clip,keepaspectratio]{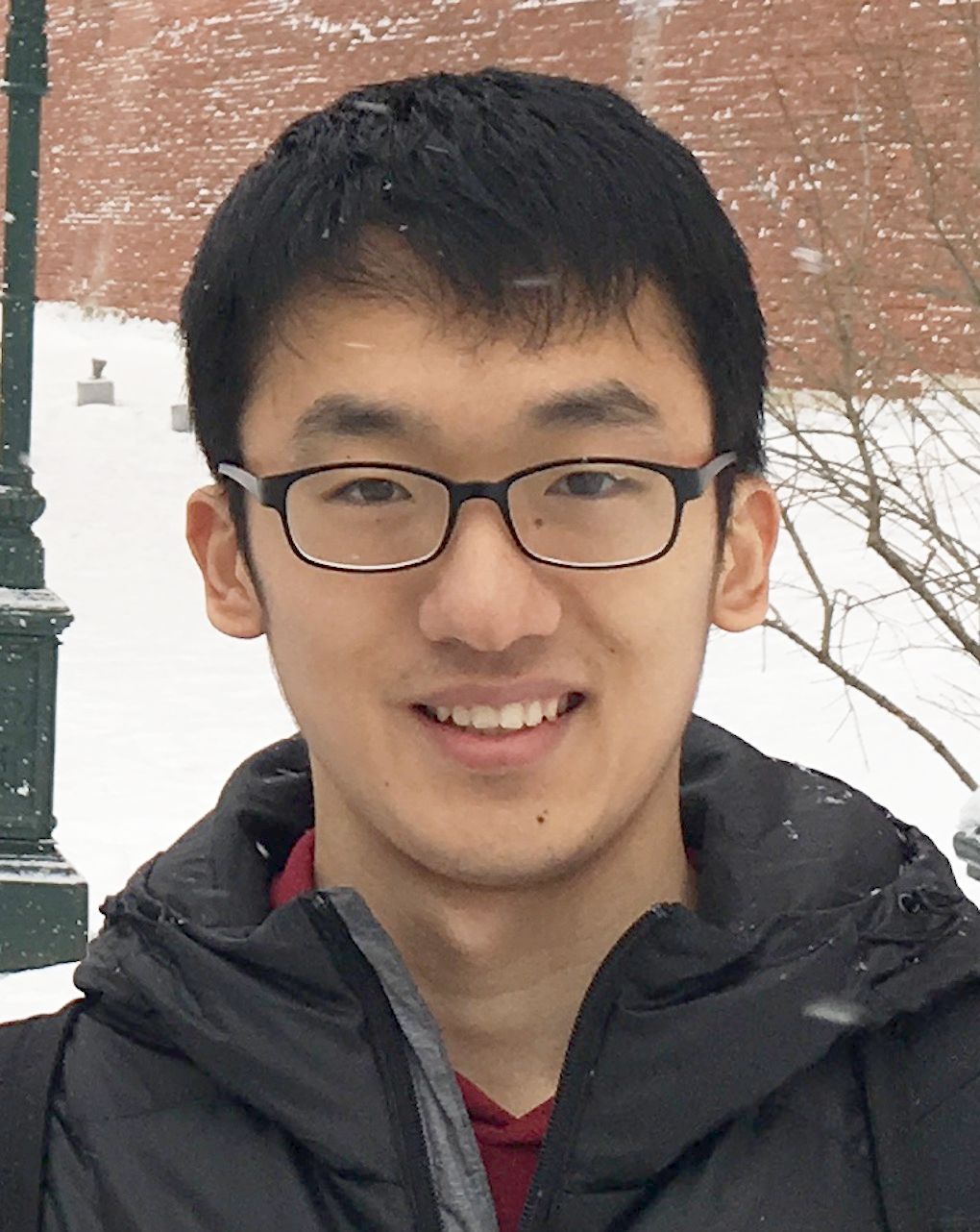}}]
{Mengyang Wu} received the B.S. degree from University College London.
He is currently a PhD student in Computer Science and Engineering of Chinese University of Hong Kong.
His recent research interests include deep learning for 3D vision, scene understanding, and outdoor augmented reality.
\end{IEEEbiography}

\vspace*{-4mm}
\begin{IEEEbiography}[{\includegraphics[width=1in,height=1.25in,clip,keepaspectratio]{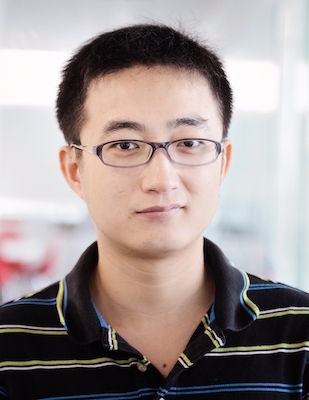}}]
{Wei Zeng} is currently an associate professor at Shenzhen Institute of Advanced Technology, Chinese Academy of Sciences.
He received his bachelor's degree (2011) and Ph.D. (2015), both in computer science, from Nanyang Technological University.
He served as co-chairs for PacificVis Poster, and program committee members for various research conferences, including IEEE Visualization, ChinaVis, IVAPP.
His recent research interests include data visualization and AR/VR, and their applications in smart city.
\end{IEEEbiography}

\vspace*{-7.5mm}
\begin{IEEEbiography}[{\includegraphics[width=1.0in,height=1.25in,clip,keepaspectratio]{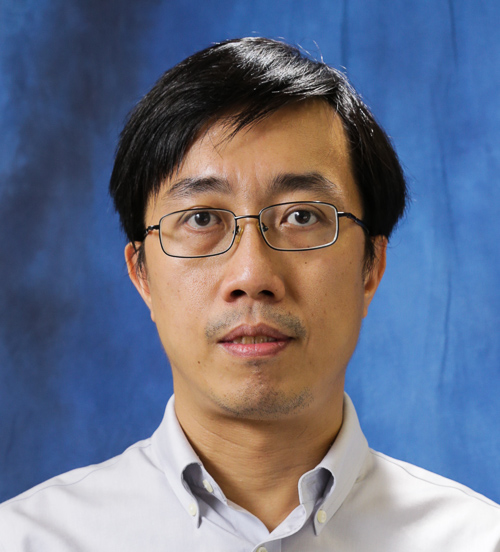}}]{Chi-Wing Fu} is currently an associate professor in the Chinese University of Hong Kong.  He served as the co-chair of SIGGRAPH ASIA 2016's Technical Brief and Poster program, associate editor of IEEE Computer Graphics \& Applications and Computer Graphics Forum, panel member in SIGGRAPH 2019 Doctoral Consortium, and program committee members in various research conferences, including SIGGRAPH Asia Technical Brief, SIGGRAPH Asia Emerging tech., IEEE visualization, CVPR, IEEE VR, VRST, Pacific Graphics, GMP, etc.  His recent research interests include computation fabrication, point cloud processing, 3D computer vision, user interaction, and data visualization.
\end{IEEEbiography}

\vfill

\end{document}